\documentclass[conference]{IEEEtran}
\usepackage{array}
\usepackage{cite}
\usepackage{hyperref}
\ifCLASSINFOpdf
   \usepackage[pdftex]{graphicx}
   \graphicspath{{Figures}}
   \DeclareGraphicsExtensions{.pdf,.jpeg,.png}
\else
\fi

\usepackage[cmex10]{amsmath}
\usepackage{flushend}
\usepackage{algorithmic}
\usepackage{algorithm}
\def\infinity{\rotatebox{90}{8}}

\usepackage{color}
\usepackage{graphicx}
\usepackage{todonotes}

\newcommand{\paragraphX}[1]{\vskip 4pt \noindent \textit{#1} \hskip .05in}

\begin{document}

\title{\textsc{\huge They Might NOT Be Giants}\\Crafting Black-Box Adversarial Examples with Fewer Queries Using Particle Swarm Optimization}

\author{\IEEEauthorblockN{Rayan Mosli\IEEEauthorrefmark{1},
Matthew Wright\IEEEauthorrefmark{2},
Bo Yuan\IEEEauthorrefmark{3}, 
Yin Pan\IEEEauthorrefmark{4}}
\IEEEauthorblockA{\IEEEauthorrefmark{1}\IEEEauthorrefmark{2}\IEEEauthorrefmark{3}\IEEEauthorrefmark{4}Golisano College of Computing and Information Sciences\\
Rochester Institute of Technology,
Rochester, New York 14623\\ Email: \{rhm6501, matthew.wright, bo.yuan, yin.pan\}@rit.edu}
\IEEEauthorblockA{\IEEEauthorrefmark{1}Faculty of Computing and Information Technology\\King Abdul-Aziz University, Jeddah, Saudi Arabia}
}

\maketitle

\begin{abstract}
Machine learning models have been found to be susceptible to adversarial examples that are often indistinguishable from the original inputs. These adversarial examples are created by applying adversarial perturbations to input samples, which would cause them to be misclassified by the target models. Attacks that search and apply the perturbations to create adversarial examples are performed in both white-box and black-box settings, depending on the information available to the attacker about the target. For black-box attacks, the only capability available to the attacker is the ability to query the target with specially crafted inputs and observing the labels returned by the model. Current black-box attacks either have low success rates, requires a high number of queries, or produce adversarial examples that are easily distinguishable from their sources. In this paper, we present AdversarialPSO, a black-box attack that uses fewer queries to create adversarial examples with high success rates. AdversarialPSO is based on the evolutionary search algorithm Particle Swarm Optimization, a population-based gradient-free optimization algorithm. It is flexible in balancing the number of queries submitted to the target vs the quality of imperceptible adversarial examples. The attack has been evaluated using the image classification benchmark datasets CIFAR-10, MNIST, and Imagenet, achieving success rates of 99.6\%, 96.3\%, and 82.0\%, respectively, while submitting substantially fewer queries than the state-of-the-art. We also present a black-box method for isolating salient features used by models when making classifications. This method, called Swarms with Individual Search Spaces or SWISS, creates adversarial examples by finding and modifying the most important features in the input. The purpose of these two attacks is to help evaluate the robustness of machine learning models and to encourage the exploration of much-needed defenses. 
\end{abstract}

\section{Introduction}
Deep learning (DL) is being used to solve a wide variety of problems in many different domains, such as image classification~\cite{Simonyan:2015}, malware detection~\cite{Raff:2018}, speech recognition~\cite{Zhang:2017}, and medicine~\cite{Carneiro:2017}. Despite state-of-the-art performances, DL models have been shown to suffer from a general flaw that makes them vulnerable to external attack. Adversaries can cause models to misclassify inputs by applying small perturbations to samples at test time~\cite{Szegedy:2014}. These \textit{adversarial examples} have even been successfully demonstrated against real-world black-box targets, where adversaries would perform remote queries on the classifier to develop and test their attacks~\cite{Papernot:2017}. The possibility of such attacks poses a significant risk to any ML application, especially in security-critical settings.

Existing attacks, both black-box and white-box, rely on model gradients to craft adversarial examples~\cite{Carlini:2017}~\cite{Goodfellow:2015}~\cite{Papernot:2016}. This requires internal knowledge of target models, which is available in a white-box attack but not a black-box one. To overcome this requirement, black-box attacks first train a surrogate model that approximates the target's decision boundary~\cite{Papernot:2017}. The surrogate is then used as a white-box model on which adversarial examples are crafted. If the surrogate sufficiently approximates the target, adversarial examples crafted on the surrogate transfer to the black-box model, where they are subsequently misclassified.

Existing approaches, however, are currently either inefficient or ineffective. The most effective existing technique by Chen et al.~\cite{Chen:2017} requires \textit{hundreds of thousands of queries} to the model, which takes time and could be easily detected. On the other hand, the black-box attack by Carlini and Wagner~\cite{Carlini:2017} uses a reasonable number of queries but only succeeds to produce an effective adversarial example between 5\% and 33\% while requiring much more computational effort. Both of these approaches rely on gradient-based optimization, which may be prone to get stuck in local minima. Gradient-based optimization is generally very effective in the white-box context, since the adversary can explore the model at leisure. In the black-box setting, however, the limited number of queries makes these approaches harder to apply.

In this paper, we examine how an adversary could overcome the limitations of gradient-based approaches to create a more realistic and thus more dangerous attack. In particular, we propose the use of Particle Swarm Optimization (PSO)---a gradient-free optimization technique---to craft adversarial examples. PSO maintains a population of candidate solutions called particles. Each particle moves in the search space seeking better solutions to the problem based on a fitness function that we have designed for the problem of finding adversarial examples. 

In our attack, called AdversarialPSO, we specify that particles move by making small perturbations to the input image that are virtually imperceptible to a human observer. PSO has been shown~\cite{Yuhui:1999} to quickly converge on good (though not globally optimal) solutions, making it very suitable for finding adversarial examples in the black-box setting, as it can identify sufficiently good examples with few queries. We demonstrate that this approach does indeed make for an effective algorithm by using three image classification datasets---MNIST, CIFAR-10, and Imagenet---and showing that PSO uses far fewer queries to the target model than prior techniques to generate adversarial examples with high success rates and small distances from the original images.    

In practice, the adversary may be constrained to making fewer queries or, alternatively, be able to make more queries and want to improve the quality of the images further. To address this need, we also explore a method that better enables a trade-off between the number of queries and image quality. This approach utilizes some aspects of PSO to determine the location of the most prominent features used by the target model when making a classification to find adversarial examples subsequently. Previous work depends on model gradients to determine salient features. With this technique, however, the adversary can determine the important features in a black-box setting without the knowledge of model internals. We call this attack Swarms With Individual Search Spaces or SWISS. SWISS restricts the particles to specific regions of the search space, where they perturb the whole area 
to determine how much the model's classification will be affected and isolate the features responsible for the model's incorrect predictions.
Since each step of localization focuses on improving progressively smaller regions of the image, the quality of the image gradually improves with more effort, leading to a natural trade-off between the effort involved and the quality of the outputs.

\paragraphX{Contributions.} In summary, our contributions are:

\begin{itemize}
    \item We present AdversarialPSO, a gradient-free black-box attack with controllable trade-offs between the number of queries and the quality of adversarial examples. 
    \item We demonstrate the effectiveness of the AdversarialPSO attack on both low-dimensional and high-dimensional datasets by empirically evaluating the attack on the MNIST, CIFAR-10, and Imagenet datasets. We show that AdversarialPSO produces adversarial examples comparable to the state-of-the-art but with significantly fewer queries and much lower computational overhead. 
    \item To isolate salient features in a black-box setting, we present the SWISS attack, an iterative localization process that operates on the same PSO infrastructure used by AdversarialPSO. The SWISS attack provides an alternative to creating black-box adversarial examples, the quality of which could be controlled by trading quality for the number of submitted queries.
\end{itemize}

\section{Related Work}
In the following section, we include related work in both white-box and black-box settings. 

\subsection{White-box Attacks}
Szegedy et. al were the first to discuss the properties of neural networks that make adversarial attacks possible~\cite{Szegedy:2014}. They show that imperceptible non-random perturbations when made to a test image can cause a DL model to misclassify the image. The attack was demonstrated using box-constrained L-BFGS that calculates the perturbations necessary to get an image misclassified. The authors also discuss the transferability of adversarial examples from one model to another, even when the models have different architectures or when they are trained using different subsets of the training data.

An explanation to why DL models are susceptible to adversarial examples was presented by Goodfellow et al.~\cite{Goodfellow:2015}, who argue that the linearity of neural networks is what leads to their sensitivity to small and directed changes in the input. They also present the Fast Gradient Sign Method (FGSM), which calculates the perturbations that transform inputs to adversarial examples. FGSM determines the direction of the perturbations according to the model gradients with respect to the input and adds minuscule values in that direction. Kurakin et al.~\cite{Kurakin:2016} extend this approach by introducing an iterative variant of FGSM that takes several smaller steps instead of one relatively larger step. The authors also evaluate the persistence of both FGSM and its iterative variant in the physical world. To do so, they print out adversarial examples and classify the images by feeding them to a model through a camera. They found that adversarial attacks are possible even in the physical world.

Papernot et al. take a different approach to craft adversarial examples~\cite{Papernot:2016}. Instead of taking multiple small steps, they construct a saliency map that maintains relevant input features with high impact on model outputs. They utilize the saliency map to perturb specific features and create adversarial examples. This approach allows the adversarial example to be crafted towards a target label specified by the attacker. In a later paper, Papernot et al. extended the techniques of both Goodfellow et al. and Papernot et al. to launch black-box attacks against remotely hosted targets~\cite{Papernot:2017}. As both attacks require knowledge of model internals---information that is not available in a black-box setting---the authors resort to a local white-box surrogate that approximates the black-box target. The surrogate is trained using the Jacobian-based Dataset Augmentation method, which expands the training set used to train the surrogate with data points that allow the surrogate to closely approximate the target's decision boundary. 

Another approach was employed by Carlini and Wagner~\cite{Carlini:2017}, in which they formulate their attack as an optimization problem. They generate adversarial examples by iteratively performing $minimize\: \mathcal{D}(x,x+\delta)$ where $\mathcal{D}$ is either an $L_{0}$, $L_{2}$, or $L_{\infinity}$ norm. The attack finds the minimum distance required to generate an adversarial example, similar to that presented by Szegedy et al.\cite{Szegedy:2014}. This is performed by reducing a set of all pixels in an image to a subset of pixels that are allowed to be changed. The attack defeats the defensive distillation approach of Papernot et al.~\cite{Papernot:2015b}.

\subsection{Black-box Attacks}
In black-box attacks, the attackers have no knowledge of the target's internals and only have the ability to query the target with inputs of their choosing. Target models are assumed to return confidence scores with each classification, which are then used in creating the inputs for subsequent queries. These inputs are specially crafted to gradually lead the attackers to generating samples that are misclassified by the model.  

To launch black-box attacks, Chen et al. propose ZOO~\cite{Chen:2017}, a method to estimate model gradients using only the model inputs and the corresponding confidence scores provided by the model. The approach uses a finite difference method that evaluates image coordinates after adding a small perturbation to estimate the direction of the gradient for each coordinate. However, as examining each and every coordinate does require a large number of evaluations, the authors propose the use of stochastic coordinate descent and attack-space dimension reduction to reduce the number of evaluations needed to approximate the gradients. Small perturbations are added in the direction of the gradient, which as shown in the FGSM attack, is sufficient to procure an adversarial example from the input. Although it can successfully create adversarial examples that are indistinguishable from the inputs, the ZOO attack does require a large number of queries to do so. The number of queries submitted by the attack depends on the dimensionality of the inputs, where it would increase with higher dimensions. Requiring a large number of queries may not be practical when launching real-world attacks as target models can easily monitor query traffic and block access for sources that are deemed suspicious. 

By utilizing Differential Evolution (DE), Su et al. show that some test samples can be misclassified by changing a single pixel~\cite{Su:2017}. Similar to the PSO algorithm used in this paper, DE is a population-based algorithm that maintains and manipulates a set of candidate solutions until an acceptable outcome is found. The objective of the one-pixel attack is to better understand the geometry of the adversarial space and the proximity of some adversarial examples to their respective inputs. The attack does not achieve high success rates due to the tight constraints used in the study. 

Another population-based black-box attack is GenAttack \cite{Alzantot:2018}, which uses Genetic Algorithm (GA) to craft adversarial examples. This attack iteratively performs the three genetic functions \textit{selection}, \textit{crossover}, and \textit{mutation}, where \textit{selection} extracts the most fit candidates in a population, \textit{crossover} produces a child from two parents, and \textit{mutation} encodes diversity to the population by applying small random perturbations. The authors propose two heuristics to reduce the number of queries used by GenAttack, namely dimensionality reduction and adaptive parameter scaling. As we show in Section IV, the AdversarialPSO attack we are proposing can create better or comparable adversarial examples than those created by GenAttack, but by using fewer queries. 

\paragraphX{Limitations.} Prior black-box attacks using gradient-based optimization to craft adversarial examples fall short for several reasons. First, gradients can get trapped in local minima, which can make the attack fail. Using many queries enables the adversary to learn enough about the model to avoid these minima, but at great cost in querying time and stealth, as we see with ZOO. Furthermore, calculating the gradient with respect to the inputs through backpropagation is resource intensive and time consuming. It requires specialized hardware such as GPUs to craft adversarial examples in an acceptable time frame. Finally, as previously mentioned, black-box attacks using gradients calculated from a local model require the intermediate step of training a surrogate, which adds to the overhead of the attack but does not guarantee the successful transfer of the adversarial example to the target model.

\section{Particle Swarm Optimization}
In this section, we provide an overview of the PSO algorithm and describe how we use it to generate adversarial examples against image classification models. 
\subsection{Conventional PSO}
Kennedy and Eberhart first proposed PSO as a model to simulate how flocks of birds forage for food~\cite{Kennedy:1995}. It has since been adapted to address a multitude of problems, such as text feature selection~\cite{Lu:2015}, grid job scheduling~\cite{Izakian:2009}, and optimizing generation of electricity~\cite{Gaing:2003}. The algorithm works by dispersing particles in a search space and moving them until a solution is found. The search space is assumed to be $d$-dimensional, where the position of each particle $i$ is a $d$-dimensional vector  $\textit{X}_{i}=(\textit{x}_{i,1},\textit{x}_{i,2},\textit{x}_{i,3},\dots,\textit{x}_{i,d})$.
The position of each particle is updated according to a velocity vector $\textit{V}_{i}$ where $\textit{V}_{i}=(\textit{v}_{i,1},\textit{v}_{i,2},\textit{v}_{i,3},\dots,\textit{v}_{i,d})$. In each time-step or iteration, denoted as \textit{t} the velocity vector is used to update the particle's next position, calculated as:
\begin{equation}\label{eq:pos}
x_{i}(t+1)=x_{i}(t) + v_{i}(t+1)
\end{equation}
\begin{equation}\label{eq:vel}
v_{i}(t+1)=wv_{t}+c_{1}r_{1}(p_{g}-x_{i}(t))+c_{2}r_{2}(p_{i}-x_{i}(t))
\end{equation}
Equation~\ref{eq:vel} contains three terms. The first term controls how much influence the current velocity has when calculating the next velocity and is constrained with the \textit{inertia} weight $w$. The second term is referred to as \textit{exploration}, as it allows particles to explore further regions in the search space in the direction of the best position found by the swarm. The best position within the swarm is denoted as $p_{g}$ and is weighted with the constant $c_{1}$. Additionally, the uniformly distributed random number $r_{1}$ is calculated in each iteration to encode randomness in the search process. Similarly, $c_{2}$ and $r_{2}$ are used to respectively weight and randomize the \textit{exploitation} portion of the search, which is performed to explore regions in the vicinity of the particle. The third term is referred to as \textit{exploitation}, and it is influenced by the best position found by each individual particle, denoted as $p_{i}$.

Early implementations of PSO assigned a fixed value to $w$. Shi and Eberhart, however, found that linearly decreasing \textit{inertia} weight was found to improve PSO performance~\cite{Yuhui:1999}. In this method, we define fixed values $w_{start}$ and $w_{end}$ together with a maximum number of iterations $t_{max}$. In each iteration, $w$ is calculated as following:
\begin{equation}\label{eq:w}
w=w_{end}+(w_{start}-w_{end})(1-\frac{t}{t_{max}})
\end{equation} 

Clerc and Kennedy further propose the use of a \textit{constriction factor} $k$ to avoid premature convergence~\cite{Clerc:2002} . PSO implementations that use $k$ often discard the \textit{inertia} weight seen in Eq.~\ref{eq:vel} to produce the following:
\begin{equation}
v_{i}(t+1)=k[v_{t}+c_{1}r_{1}(p_{g}-x_{i}(t))+c_{2}r_{2}(p_{i}-x_{i}(t))]
\end{equation}
To reap the benefits of both the \textit{inertia} weight and the \textit{constriction factor}, Lu et. al propose utilizing both variables when performing PSO~\cite{Lu:2015}, which can be done synchronously and asynchronously as seen in Equation~\ref{eq:kw-syn} and Equation~\ref{eq:kw-asyn}, respectively.
\begin{equation}\label{eq:kw-syn}
v_{i}(t+1)=k[wv_{t}+c_{1}r_{1}(p_{g}-x_{i}(t))+c_{2}r_{2}(p_{i}-x_{i}(t))]
\end{equation} 
\begin{equation}\label{eq:kw-asyn}
\left\{\begin{array}{r}v_{i}
(t+1)=wv_{t}+c_{1}r_{1}(p_{g}-x_{i}(t))+c_{2}r_{2}(p_{i}-x_{i}(t))\: \\
if \: t < \frac{t_{max}}{2}\\\\v_{i}
(t+1)=k[v_{t}+c_{1}r_{1}(p_{g}-x_{i}(t))+c_{2}r_{2}(p_{i}-x_{i}(t))]\:\\ 
if \: t \ge \frac{t_{max}}{2} \end{array}\right.
\end{equation} 
\vskip 0.5cm

For long running PSO processes with a large $t_{max}$, there is a high chance the \textit{constriction factor} portion of Equation~\ref{eq:kw-asyn} will not be reached. Hence, using this method would be equivalent to using Equation~\ref{eq:vel} to calculate $v$ if the number of iterations does not exceed $t_{max}/2$. 

\subsection{Adversarial PSO}
Among the many applications for PSO, we show in this paper that it can also be used to craft adversarial examples. Shi and Eberhart~\cite{Yuhui:1999} found that PSO is quick to converge on a solution and scales well to large dimensions, at the cost of slower convergence to global optima. This would make PSO an excellent fit for finding adversarial examples in the black-box setting, as it suggests that it can identify sufficiently good examples with few queries. 

To adapt PSO to the problem of creating adversarial examples, we define a fitness function that measures the change in model output when perturbations are added to the input. In both targeted and untargeted attacks, the fitness function simply measures how much the model's confidence in the target label rises or drops, respectively. When performing untargeted attacks, the fitness for each candidate solution is the confidence drop in the original class predicted by the model. In targeted attacks however, the fitness is the rise in confidence in the desired class.

We introduce an MSE $L_{2}$ penalty term to the fitness function to encourage smaller perturbations when synthesizing adversarial examples. Let the original image be $x$, the current adversarial image be $x'$, and $p_0$ and $p_1$ be the model's confidence in predicting the label for $x$ and $x'$, respectively. Then, the fitness function we use to generate adversarial examples is:
\begin{equation}\label{eq:fitness}
Fitness=\left|(p_{0}-p_{1})\right|-\frac{c}{n}\left\|x-x^{'}\right\|_{2},
\end{equation}  
where $c$ is a constant to weight the penalty term.

To further control the perturbations added to the input image, we define an upper bound value $B$ of maximum change to limit the $L_{\infty}$ distance between the adversarial image and the original image. $L_{\infty}$ measures the maximum change to any of the coordinates, where $L_{\infty} = max(|x_{1} - x_{1}'|,|x_{2}-x_{2}'|,\dots,|x_{d}-x_{d}'|)$. Essentially, we use the clip operator such that $x^{'}=\displaystyle clip(x_{i}+v_{i},x_{i}-B,x_{i}+B)$. Additionally, we also apply box constraints to maintain valid image values when adding perturbations. These constraints are applied to Equation~\ref{eq:pos} to yield:
\begin{equation}\label{eq:pos-box}
x_{i}(t+1)=clip(clip(x_{i}(t)+v_{i}(t+1),x_{i}-B,x_{i}+B),0,1)
\end{equation}
For each image, generation of adversarial examples occurs in three steps: initialization, optimization, and reduction.
\subsubsection{Initialization} 
Initialization of the swarm involves creating the particles and dispersing them across the search space. Two hyperparameters control how the swarm is initialized: the number of particles in the swarm $P$ and the change rate $m$, which determines how widely the particles are dispersed. Each particle begins with the input image $x$, but with a random subset of their coordinates perturbed. The change rate specifies how many of the total coordinates are perturbed and, as such, controls how dispersed the particles are across the search space. Although a highly dispersed swarm would have more coverage, scattering the particles far from the original image would result in adversarial examples that are quite distant from the input. Once the particles are created and dispersed, their fitness is calculated and subsequently used in the optimization step of PSO. The algorithm for the initialization step is shown in Algorithm~\ref{alg:init}.
\begin{algorithm}
\caption{Initializing the swarm}
\begin{algorithmic}[1]\label{alg:init}
\STATE \textbf{Input:} Input image $x$, penalty weight $c$, particle array $par$, number of particles $P$, change rate $m$
\STATE $d\gets$ length\:of\:$x$
\STATE $indexes\gets$\:select\:random\:$d*m$\:elements\:from\:$x$
\STATE $bestFitness\gets 0$  \# swarm-wide best
\STATE $bestPosition\gets x$
\FOR{$p\gets 0$ to $P-1$}
\STATE $par[p].position\gets x$
\FOR{$i\:\in\:indexes$}
\STATE $par[p].position_{i}\gets par[p].position_i+\epsilon$
\ENDFOR
\STATE $par[p].bestFit\gets calculateFitness(c)$
\STATE $par[p].bestPos\gets par[p].position$
\IF{$par[p].bestFit > bestFitness$}
\STATE $bestFitness\gets par[p].bestFit$
\STATE $bestPosition\gets par[p].bestPos$
\ENDIF
\ENDFOR
\RETURN $par$, $bestPosition$, $bestFitness$
\end{algorithmic}
\end{algorithm}

\subsubsection{Optimization} The optimization step of AdversarialPSO is an iterative process that moves the particles in search for better fitness. Particle positions are updated using the velocity vector, which is calculated for each particle in every iteration. Although traditionally the velocity vector is used to calculate both the direction and the step size, in our implementation, we only use it to determine the direction in which the particles are moved along each dimension. In each iteration, the step size is multiplied by a randomly generated number from a uniform distribution between 0.0 and 1.0 and is then added to the value of the current coordinate in the direction produced by the velocity. After moving the particles, their fitness is calculated using Equation~\ref{eq:fitness}. The new particle fitness is compared against the particle's best fitness to determine which particle position will be used to calculate future particle movements. At the end of each iteration, the particle with the highest fitness is compared against the best fitness achieved in the swarm as a whole (i.e, best swarm fitness), and if the particle fitness was found to be better, the swarm is updated to account for the position with the highest fitness. The process is repeated until the fitness threshold is reached or if the process exhausts the allowed number of iterations. The algorithm for the optimization step can be seen in Algorithm~\ref{alg:opt}.

\begin{algorithm}
\caption{Optimization}
\begin{algorithmic}[1]\label{alg:opt}
\STATE \textbf{Input:} Input image $x$, maximum iterations $t_{max}$, maximum change $B$, penalty weight $c$, particle array $par$, number of particles $P$, change rate $m$, stopping criteria $fit_{max}$, swarm-wide $bestFitness$, and $bestPosition$
\WHILE{$t < t_{max}$}
\IF{$bestFitness>fit_{max}$}
\RETURN $bestPosition$
\ENDIF
\FOR{$p\gets 0$ to $P$}
\STATE $v\gets calculateVelocity$
\STATE $par[p].position\gets updatePosition(B)$
\STATE $fitness\gets calculateFitness(c)$
\IF{$fitness > par[p].bestFitness$}
\STATE $par[p].bestFit\gets fitness$
\STATE $par[p].bestPos\gets par[p].position$
\ENDIF
\IF{$par[p].bestFit > bestFitness$}
\STATE $bestFitness\gets par[p].bestFit$
\STATE $bestPosition\gets par[p].bestPos$
\ENDIF
\ENDFOR
\ENDWHILE
\RETURN $bestPosition$
\end{algorithmic}
\end{algorithm}

\subsubsection{Reduction}\label{sec:reduction} This step reduces the distance between the adversarial example generated by PSO and the original image. Considering the different geometric distances between inputs and their closest adversarial example, some images might require more changes than others~\cite{Gilmer:2018}. We have thus found that setting the step-size to a small fixed value would in some situations fail to find adversarial examples, or would do so only after submitting a large number of queries. To successfully generate adversarial examples from these images, larger step-sizes must be used, which can result in overshooting the target and thus adding unnecessary perturbations. To alleviate this problem, we make two changes to the algorithm. First, rather than starting with a large step-size, we commence the search by adding small perturbations and only increase the step-size if several iterations go by with no improvement to the best fitness in the swarm. This is similar to the adaptive parameter scaling method used in GenAttack~\cite{Alzantot:2018}, but instead of decreasing the step-size, we increase it. Second, after successfully finding an adversarial example, we remove all unnecessary perturbations introduced by PSO. The reduction task is a simple iterative operation that reduces the difference between the adversarial image and input image by half until the label changes, at which point it stops and takes one step back. The reduction process does increase the total number of queries made by PSO, but this can be easily controlled by limiting the number reduction iterations. The algorithm for the reduction step can be seen in Algorithm~\ref{alg:reduce}.

\begin{algorithm}
\caption{Reducing excess perturbations from adversarial image}
\begin{algorithmic}[1]\label{alg:reduce}
\STATE \textbf{Input:} Input image $x$, Best position $bp$
\FOR{$i\gets$ indices\:of\:perturbed\:coordinates}
\IF{adversarial\:image\:label\:$==$\:target\:label}
\STATE{$\mathit{diff}\gets (x[i]-bp[i])*0.5$}
\STATE{$bp[i]\gets bp[i]+\mathit{diff} $}
\ELSE
\STATE{undo\:last\:changes}
\RETURN $bp$
\ENDIF
\ENDFOR
\RETURN $bp$
\end{algorithmic}
\end{algorithm}

\subsubsection{Baseline, Center, and Random Mutation}
In addition to the PSO algorithm we use to generate adversarial examples, we implement three low cost operations that, in some instances, improved the outcome of the search process. We found that, by randomly designating one particle as what we call a baseline particle, adversarial examples with low $L_{2}$ distances can be found with little or no reduction. In essence, the baseline particle is reverted back to the original input at the beginning of each iteration from where it takes a single step towards the swarm's best position $p_{g}$. If an adversarial example exists within a single step distance towards $p_{g}$, it can be found using the baseline particle.

Another low-cost operation we found to be beneficial is testing the fitness at the centroid of the swarm. Considering the distance penalty applied to the fitness and the common proximity of the centroid to the source image, testing the centroid can potentially yield better fitness than current particle positions. Both the center and the baseline are calculated once per iteration and as thus adds only one query each in every iteration. However, they both have the potential to produce adversarial examples with low $L_{2}$ distances, which in turn reduces the number of queries needed in the reduction phase.

Finally, we have adopted the mutation concept from genetic algorithms to encode more randomness in particle movements~\cite{Esquivel:2003}. Although we set a low probability for mutation, this nonetheless allows the exploration of regions in the search space that would have otherwise gone unexplored. To mutate a particle, we choose a random subset of coordinates according to the change rate $m$ and perturb those coordinates in a random direction. After mutating a particle, we test the fitness at the new location and update the swarm accordingly.  

\subsection{SWISS Attack}
The purpose of the SWISS attack is to isolate the salient features that influence a model's outcome. The approach we propose utilizes the PSO infrastructure to perturb different regions of the search space and observe the change in model predictions. To do so, we generate individual search spaces for each particle, where each search space is a consecutive chunk with a random length between one element of the search space and $(d/P)*2$ elements, where $d$ is the number of dimensions in the search space and $P$ is the number particles in the swarm ($d$ is typically much larger than $P$). The final particle is simply assigned all remaining elements. 

Using larger swarms in the SWISS attack would result in smaller search spaces for each particle and as such, would commence the search with higher precision. However, as some inputs require more change than others to produce adversarial examples, modifying small regions might not produce the desired output right away and would require more effort, especially in targeted attacks where a specific label is sought. To overcome this, we apply a similar heuristic as the step-size heuristic used in AdversarialPSO, where the step-size is increased if no change is observed in several iterations. Rather than increasing the step-size, however, we instead reduce the number of particles to allow for larger search spaces for each particle.

The idea behind the SWISS attack is to create a large region (in $L_{2}$ space) around the input and check if it contains adversarial examples. The region is then iteratively made smaller until it can longer be reduced. In the first iteration, the region is expanded in every direction, where each particle is responsible for expanding a subset of the dimensions and checking if the expanded region contains an adversarial example. In the next iteration, only the dimensions that contain adversarial examples are retained and divided among the particles, where the process is then repeated using the original input but with only the retained coordinates. The localization process is continued until no further localization is possible.
\begin{figure*}[!t]
\begin{tabular}{lm{1in}m{1in}m{1in}m{1in}m{1in}m{1in}}
(a)  &
\includegraphics[width=1.0in]{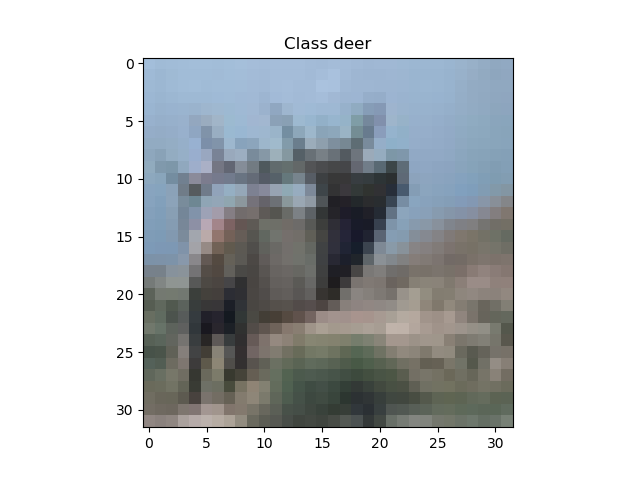} &  
\includegraphics[width=1.0in]{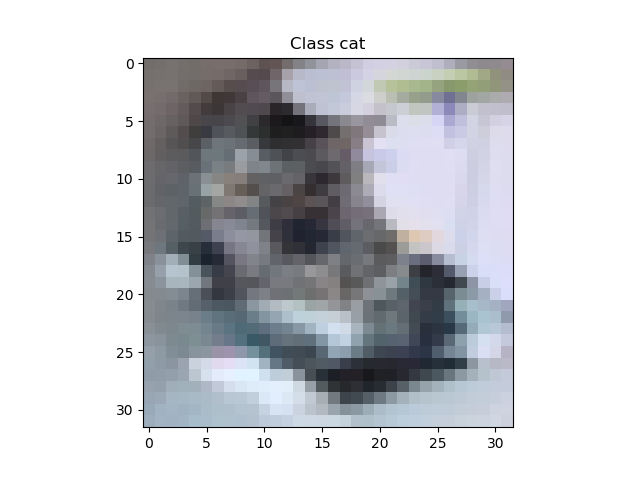} & 
\includegraphics[width=1.0in]{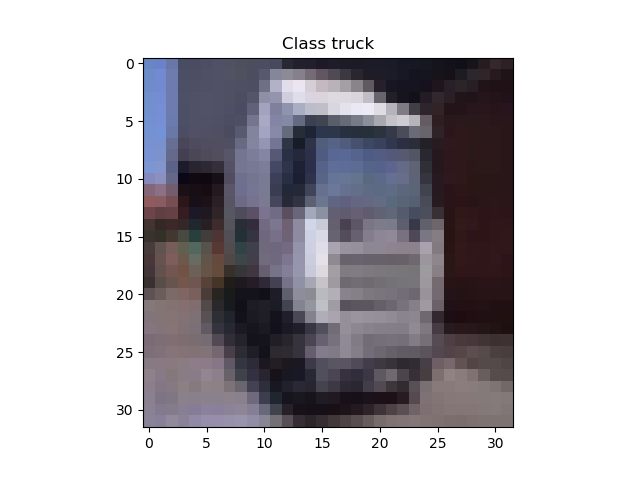} &
\includegraphics[width=1.0in]{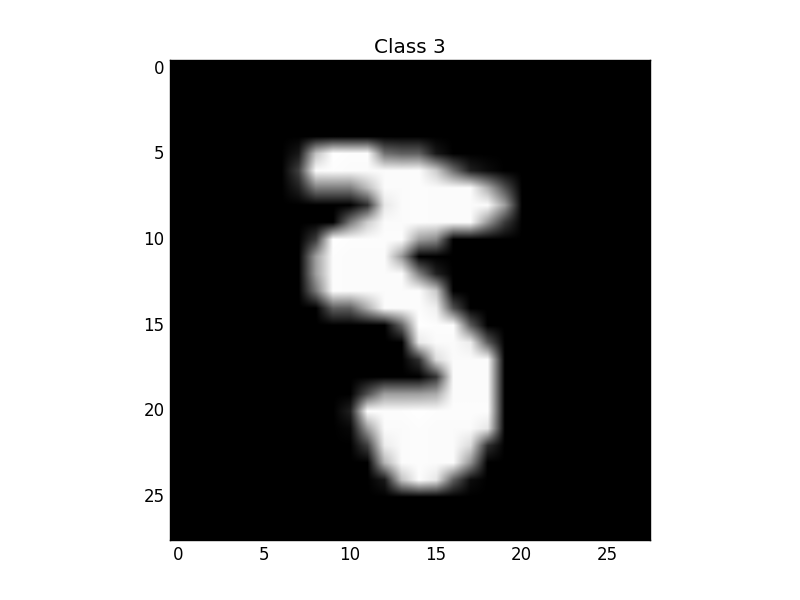} &  
\includegraphics[width=1.0in]{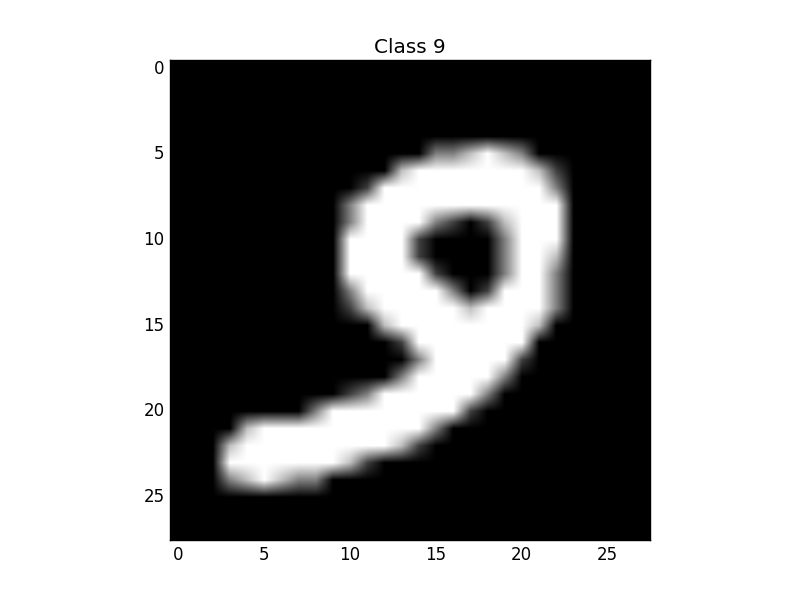} & 
\includegraphics[width=1.0in]{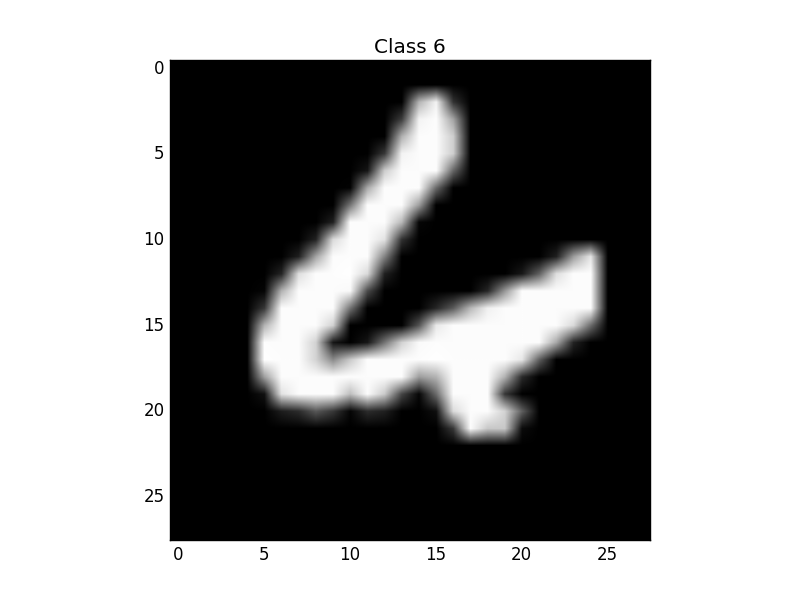} 
\\
\multicolumn{1}{c}{} & \multicolumn{1}{c}{Label: Deer} & \multicolumn{1}{c}{Label: Cat} & \multicolumn{1}{c}{Label: Truck} & \multicolumn{1}{c}{Label: 3} & \multicolumn{1}{c}{Label: 9} & \multicolumn{1}{c}{Label: 6}
 \\
(b) &
\includegraphics[width=1.0in]{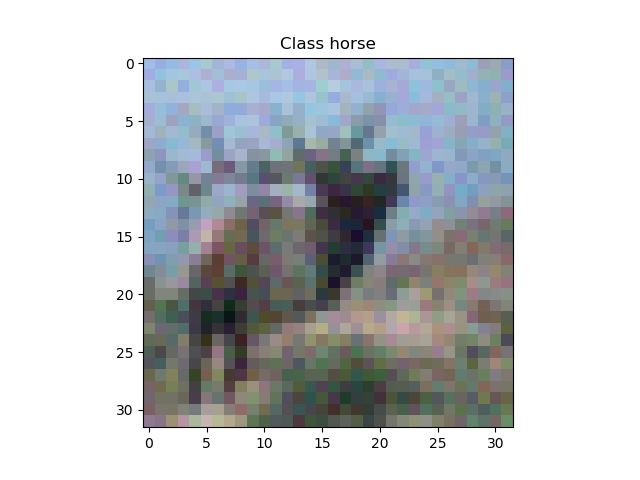} & \includegraphics[width=1.0in]{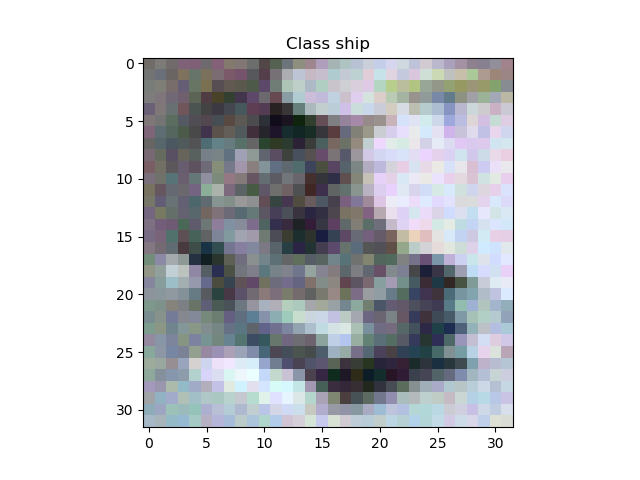} & 
\includegraphics[width=1.0in]{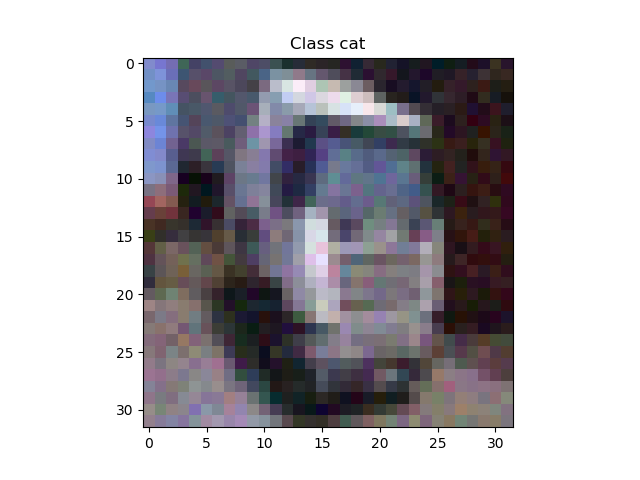} &
\includegraphics[width=1.0in]{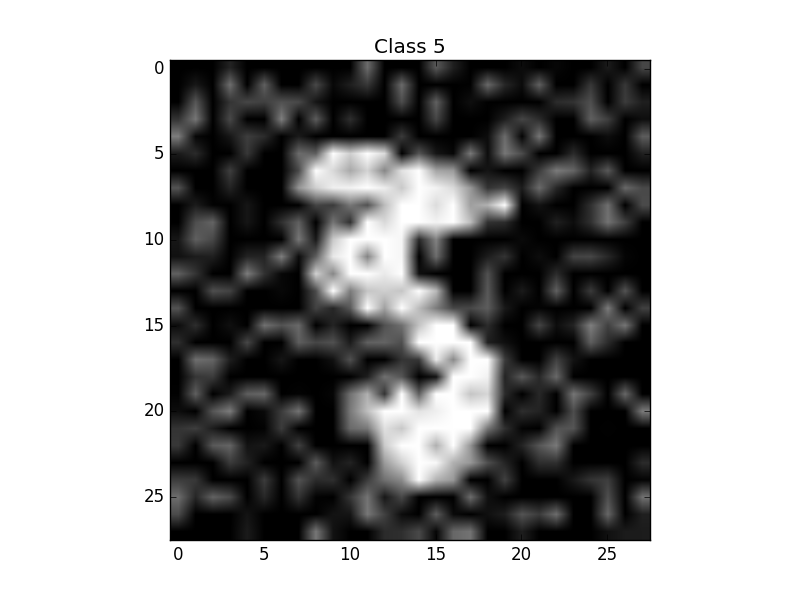} &  
\includegraphics[width=1.0in]{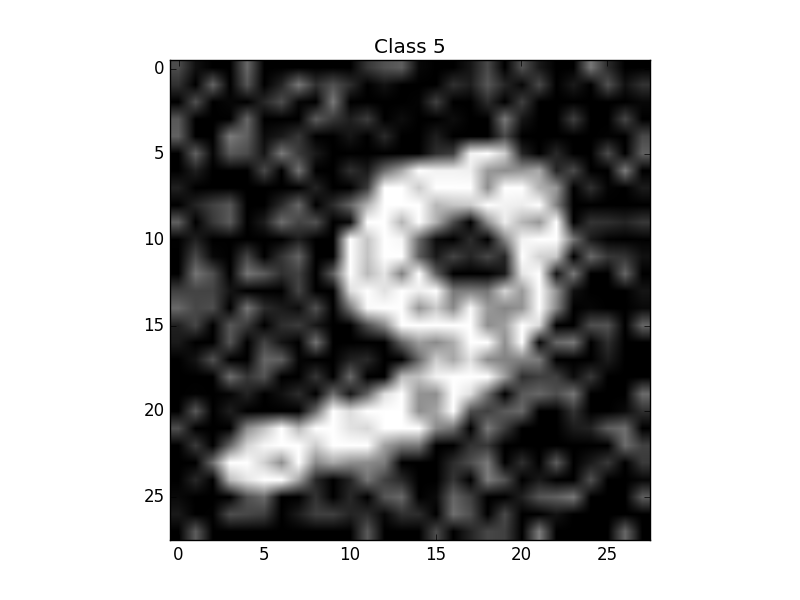} & 
\includegraphics[width=1.0in]{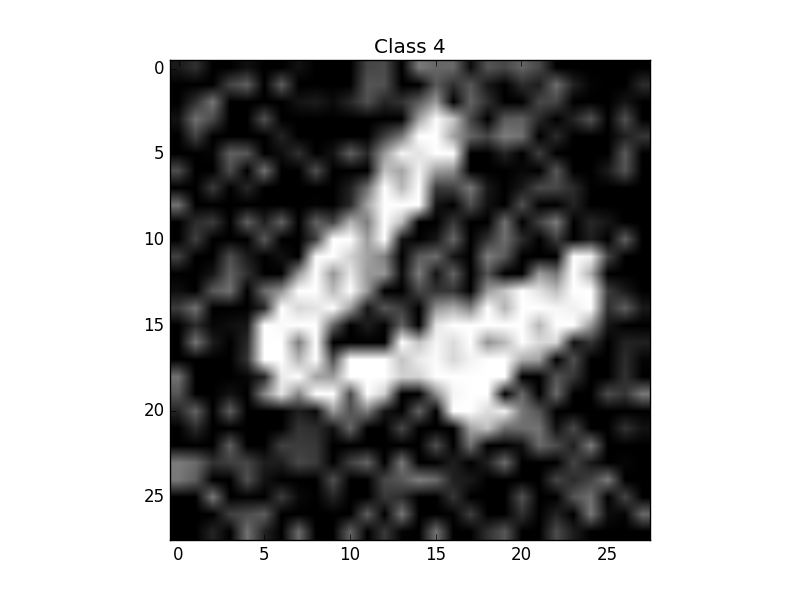} 
\\
\multicolumn{1}{c}{} & \multicolumn{1}{c}{Label: Horse} & \multicolumn{1}{c}{Label: Ship} & \multicolumn{1}{c}{Label: Cat} & \multicolumn{1}{c}{Label: 5} & \multicolumn{1}{c}{Label: 5} & \multicolumn{1}{c}{Label: 4}
 \\
(c) &
\includegraphics[width=1.0in]{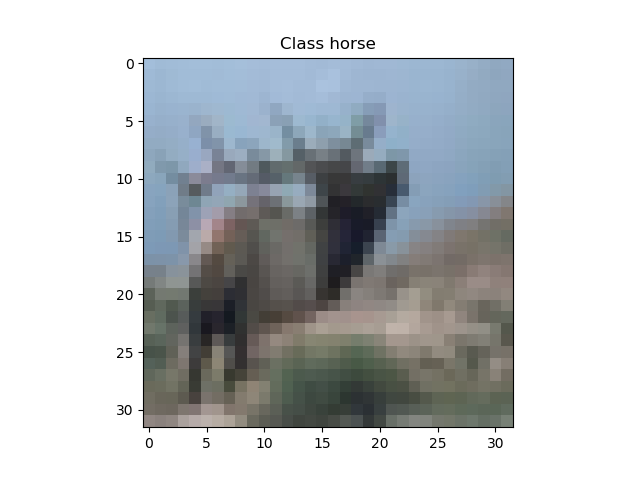} &
\includegraphics[width=1.0in]{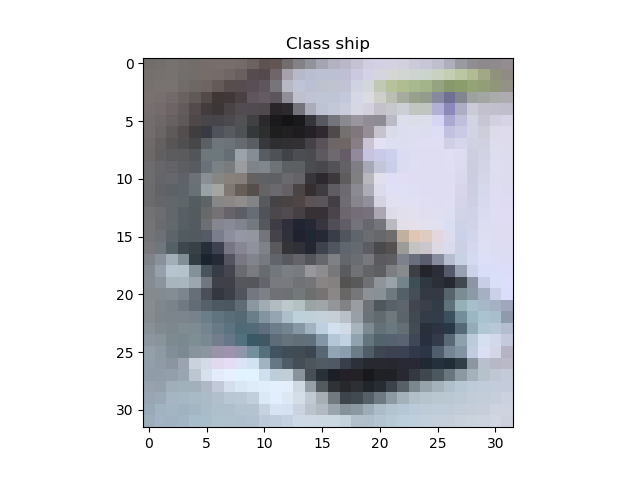} &
\includegraphics[width=1.0in]{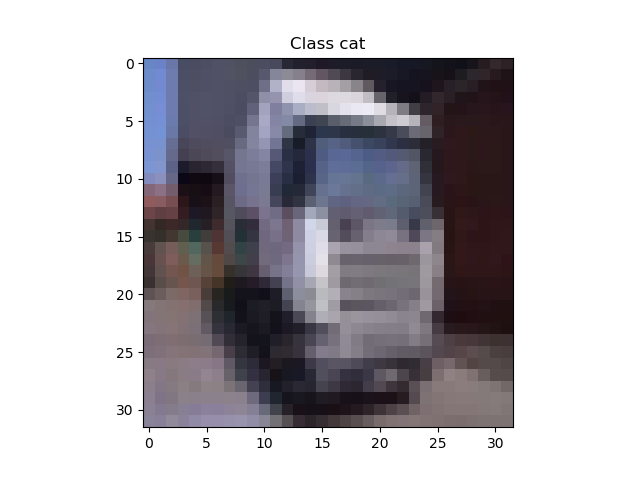} &
\includegraphics[width=1.0in]{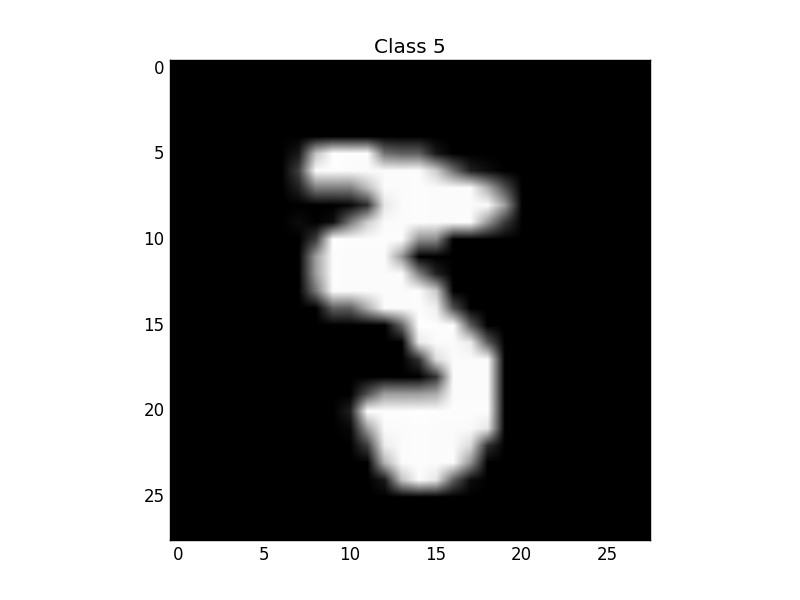} &  
\includegraphics[width=1.0in]{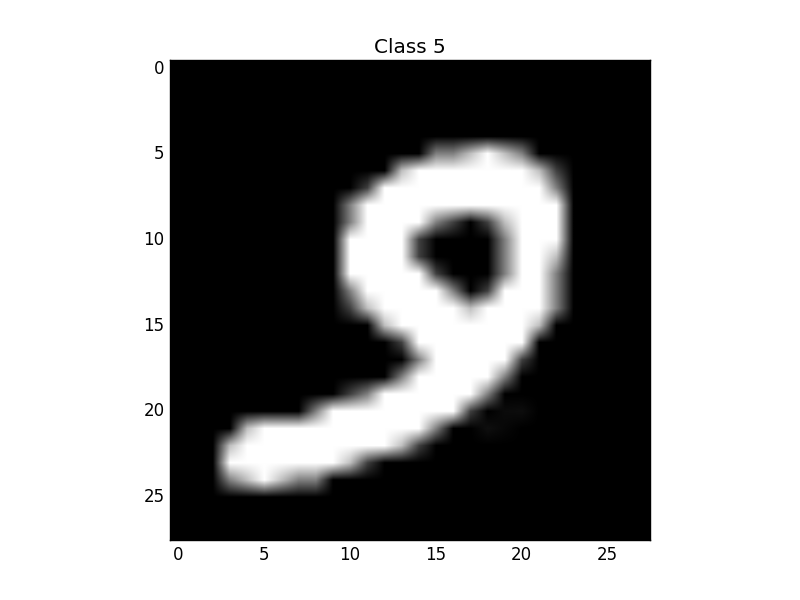} & 
\includegraphics[width=1.0in]{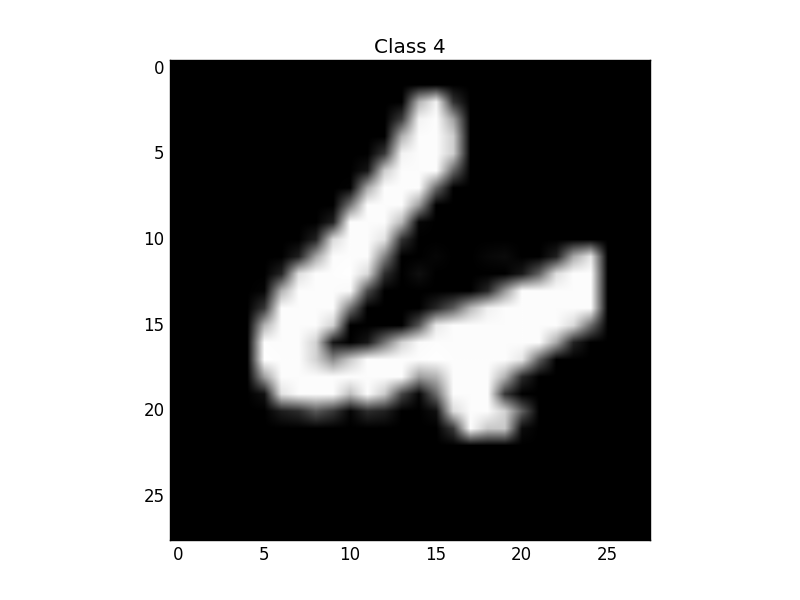}
\\
\multicolumn{1}{c}{} & \multicolumn{1}{c}{Label: Horse} & \multicolumn{1}{c}{Label: Ship} & \multicolumn{1}{c}{Label: Cat} & \multicolumn{1}{c}{Label: 5} & \multicolumn{1}{c}{Label: 5} & \multicolumn{1}{c}{Label: 4}
\end{tabular}
\caption{Untargeted AdversarialPSO attacks on CIFAR-10 and MNIST : a) Before AdversarialPSO (b) After AdversarialPSO but before reduction (c) After reduction}
\end{figure*}
\section{Evaluation}
\subsection{Setup}
To evaluate both AdversarialPSO and SWISS, we consider the following three metrics: the success rate (i.e., the ratio of successfully generated adversarial examples over the total number of samples), the average $L_{2}$ distance between input images and their adversarial counterparts, and the average number of queries needed to generate adversarial examples. We compare our results against the C\&W attack~\cite{Carlini:2017}, ZOO~\cite{Chen:2017} and GenAttack~\cite{Alzantot:2018} using the three benchmark datasets MNIST, CIFAR-10, and Imagenet. For all three datasets, we use the same models used to evaluate the three prior works~\cite{Chen:2017,Carlini:2017,Alzantot:2018};
we refer the readers to Carlini and Wagner's paper~\cite{Carlini:2017} for more details. The results we report are obtained from running both AdversarialPSO and SWISS on the first 1,000 correctly classified samples from the test sets of both MNIST and CIFAR-10 for the untargeted attacks and the first 111 correctly classified samples for the targeted attacks, which we evaluate by repeating the process for each target label. In other words, we run both the AdversarialPSO and SWISS attacks nine times for each sample, which brings the total number of attack samples to 999 per dataset. Average attack times are measured on a Lenovo ThinkPad Laptop with 24 GB RAM and an Intel Core i7-6600U 2.6 GHz CPU. The attacks are executed sequentially by moving one particle at a time with no parallel processing. 

The parameters used for the CIFAR-10 dataset in the AdversarialPSO attack are as following: 0.05 for step size, 0.05 for the $L_{\infty}$ bound, 0.1 for the probability of random mutation, 2.0 for the \textit{exploration} weight C1, 0.5 for the \textit{exploitation} weight C2, 0.05 for the change rate, 1.0 for the $L_{2}$ distance penalty, and 10 particles. For MNIST, the parameters are as following: 0.5 for the step size, 0.5 for the $L_{\infty}$ bound, 0.3 for the probability of random mutation, 2.0 for the \textit{exploration} weight C1, 2.0 for the \textit{exploitation} weight C2, 0.3 for the change rate, 1.0 for the $L_{2}$ distance penalty, and 10 particles. The value of the change rate parameter was chosen according to the results of the hyper-parameter test we describe in section \ref{sec:partest}, where we observe a negative effect with higher change rates. Therefore, for both MNIST and CIFAR-10, a small value was chosen according to their dimensionality to avoid the negative impact of changing too many coordinates. For the step-size and $L_{\infty}$ parameters, we simply chose a small value as was discussed in section \ref{sec:reduction}, to apply the step-size heuristic for samples that require larger changes without affecting samples that do not require as much change. The values of the remaining parameters were arbitrarily chosen.

For the AdversarialPSO attack, if there is no improvement to the fitness for five consecutive iterations, we double the step-size and $L_{\infty}$ bound. We do this to take advantage of the low query count of small swarms while maintaining a high effective rate and low $L_{2}$ distance. Although this only occurs for a subset of the test samples, we believe it is necessary to accommodate samples with adversarial counterparts that are further away in the search space. The downside of using this heuristic is the added queries in the reduction phase to minimize the unnecessary perturbations introduced when taking larger steps. Essentially, this heuristic favors low $L_{0}$ counts with high $L_{\infty}$ over high $L_{0}$ and low $L_{\infty}$, while maintaining balance for the $L_{2}$ distance. The $L_{0}$ metric measures the number of pixels changed between the input and adversarial examples. This translates to reaching a solution quickly by making large steps towards fewer dimensions and compensating for these steps in the reduction stage.

For SWISS attack evaluations that use more than 10 particles, if no improvement to fitness is reported in five consecutive iterations, we reduce the number particles by 10\% and we re-divide the search space among the particles. This will assign a bigger chunk of the search space to each particle in case the features needed to produce the adversarial examples were located in different chunks. By using fewer particles with larger individual search spaces, there is a higher chance that salient features would reside in the same chunk of search space. The minimum number of particles allowed in this situation is two, which would split the search space in half. In essence, running the localization from start to finish using two particles is similar to a binary search of the search space for adversarial examples. However, using just two particles from the start is not recommended, as both chunks could include important features, such that the localization loop will repeatedly return both chunks. In both attacks, if there was no improvement to the fitness for 20 consecutive iterations, we terminate the search process.  
\begin{figure*}[t]
\begin{tabular}{l m{1in}m{1in}m{1in}m{1in}m{1in}m{1in}}
(a)  &
\includegraphics[width=1.0in]{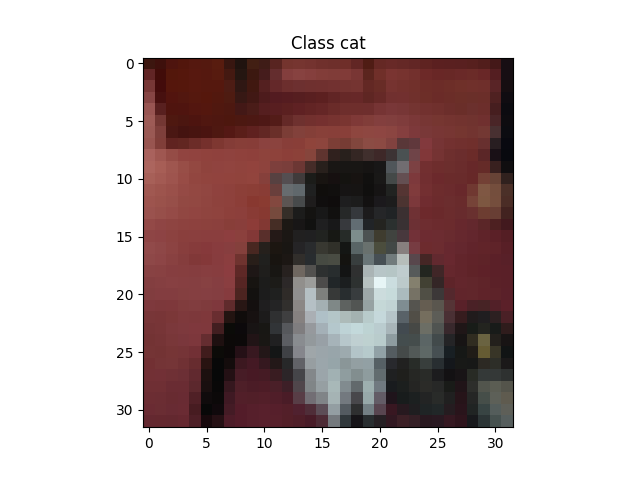} &  
\includegraphics[width=1.0in]{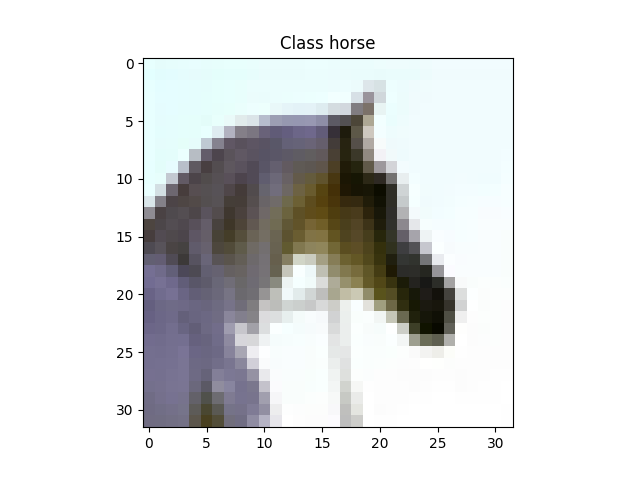} & 
\includegraphics[width=1.0in]{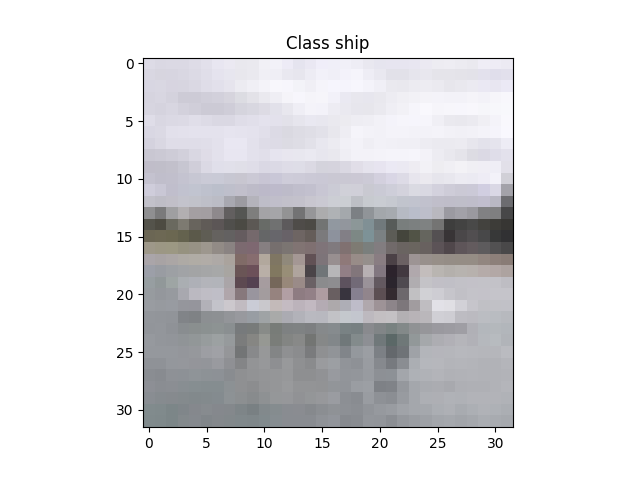} &
\includegraphics[width=1.0in]{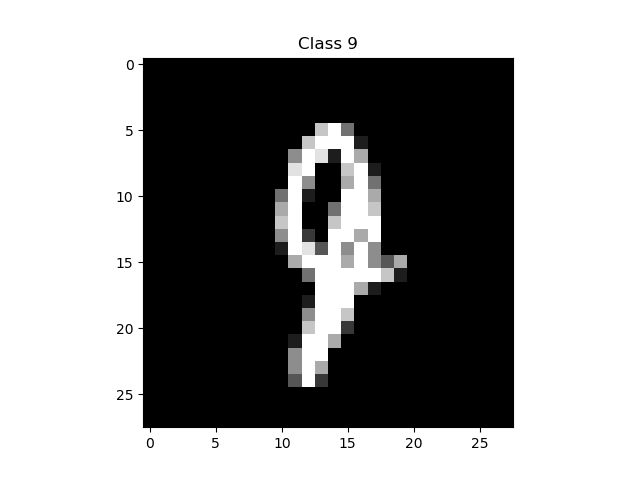} &  
\includegraphics[width=1.0in]{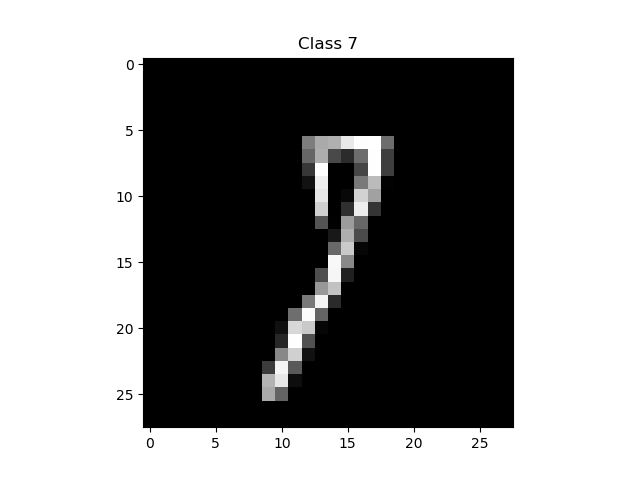} & 
\includegraphics[width=1.0in]{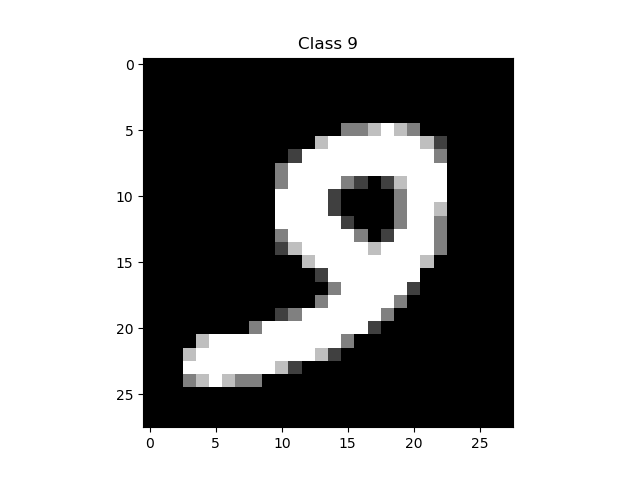} 
\\
\multicolumn{1}{c}{} & \multicolumn{1}{c}{Label: Cat} & \multicolumn{1}{c}{Label: Horse} & \multicolumn{1}{c}{Label: Ship} & \multicolumn{1}{c}{Label: 9} & \multicolumn{1}{c}{Label: 7} & \multicolumn{1}{c}{Label: 9}
\\

(b) &
\includegraphics[width=1.0in]{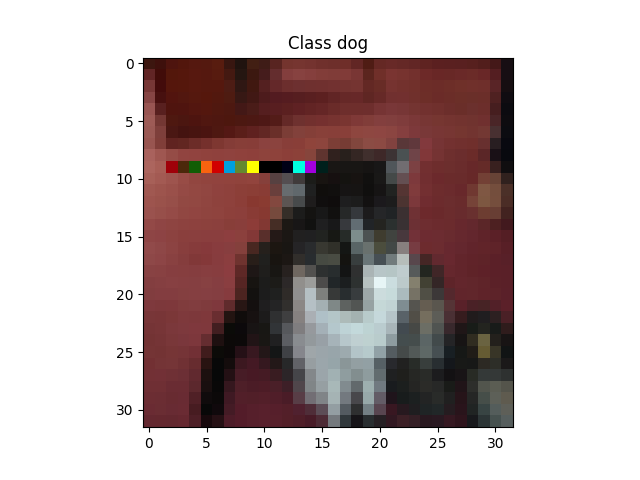} &  
\includegraphics[width=1.0in]{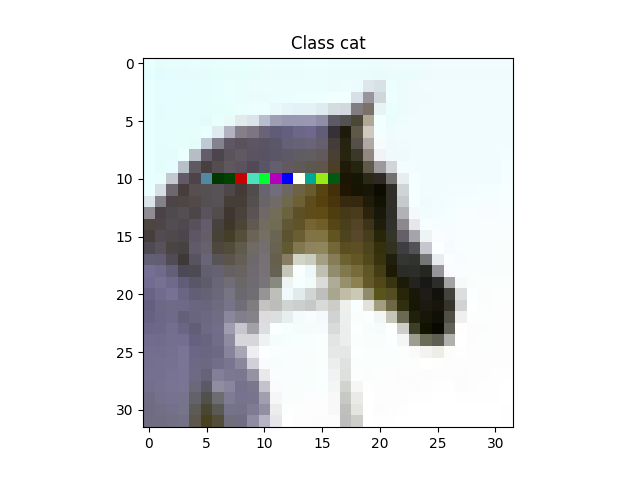} & 
\includegraphics[width=1.0in]{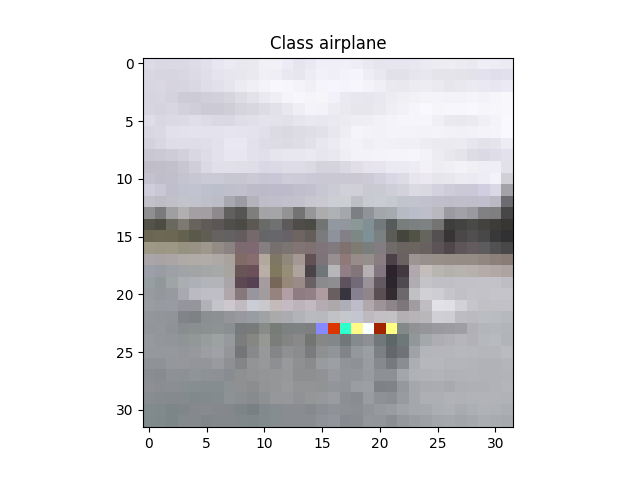} &
\includegraphics[width=1.0in]{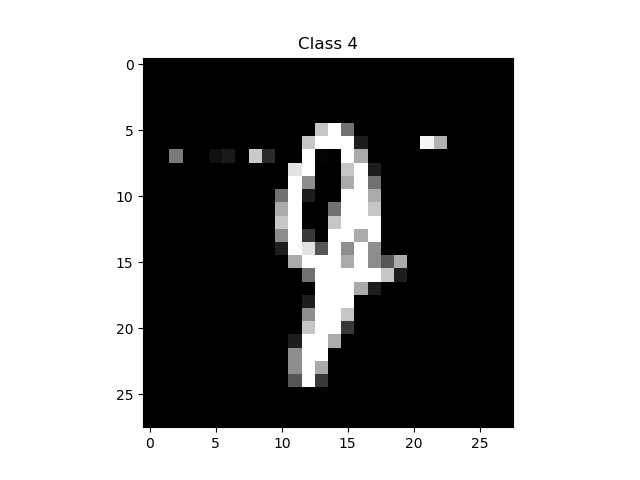} &  
\includegraphics[width=1.0in]{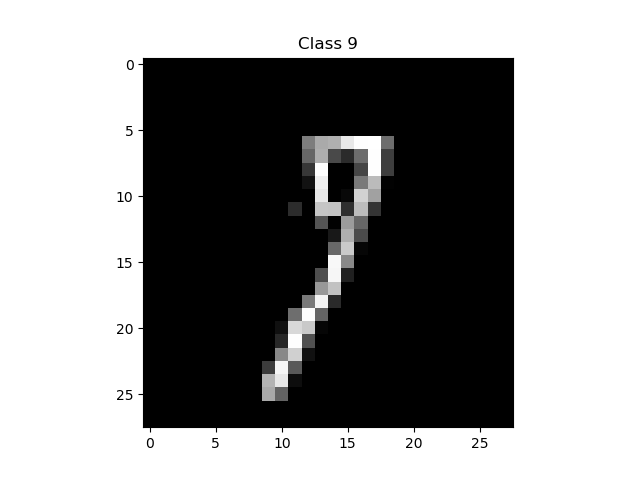} & 
\includegraphics[width=1.0in]{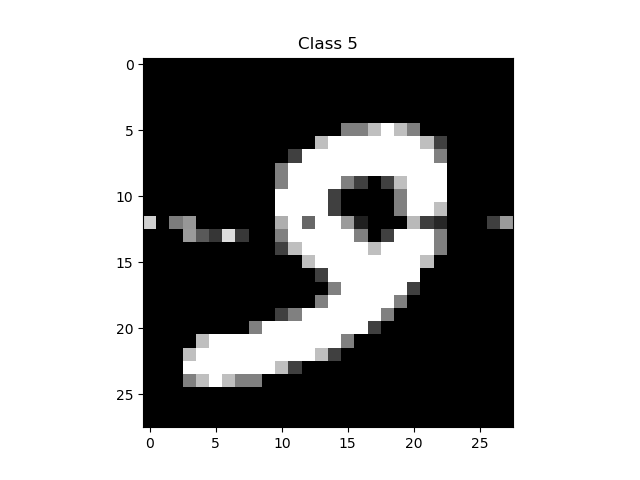} 
\\
\multicolumn{1}{c}{} & \multicolumn{1}{c}{Label: Dog} & \multicolumn{1}{c}{Label: Cat} & \multicolumn{1}{c}{Label: Airplane} & \multicolumn{1}{c}{Label: 4} & \multicolumn{1}{c}{Label: 9} & \multicolumn{1}{c}{Label: 5}
\\
(c) &
\includegraphics[width=1.0in]{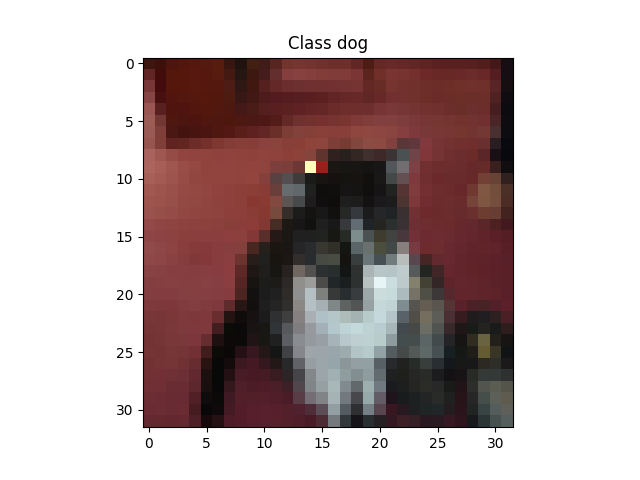} &  
\includegraphics[width=1.0in]{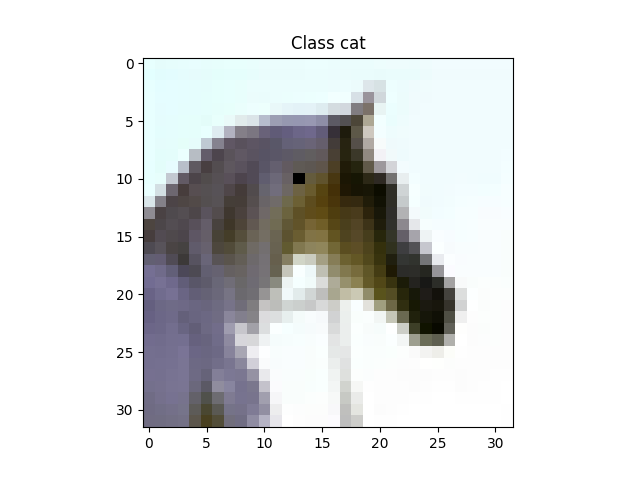} & 
\includegraphics[width=1.0in]{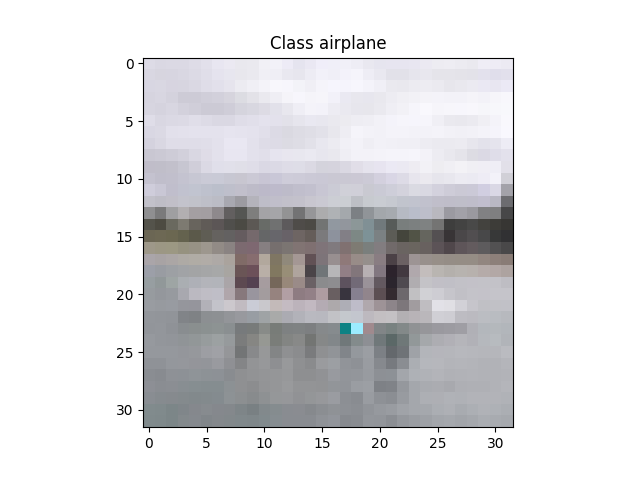} &
\includegraphics[width=1.0in]{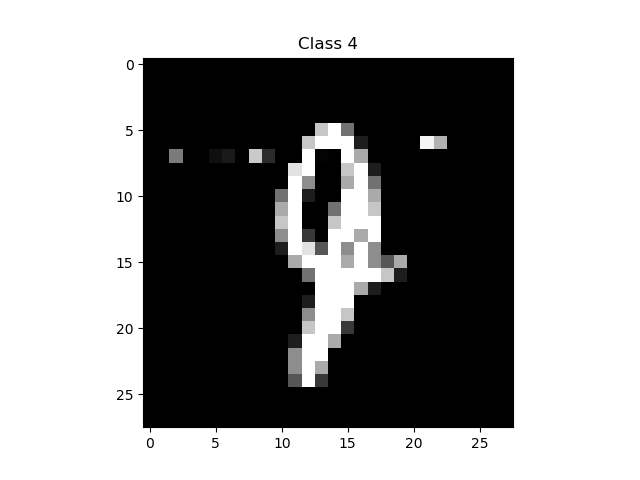} &  
\includegraphics[width=1.0in]{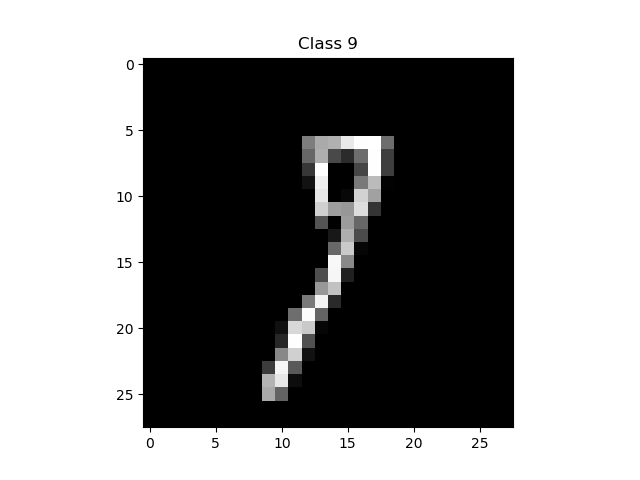} & 
\includegraphics[width=1.0in]{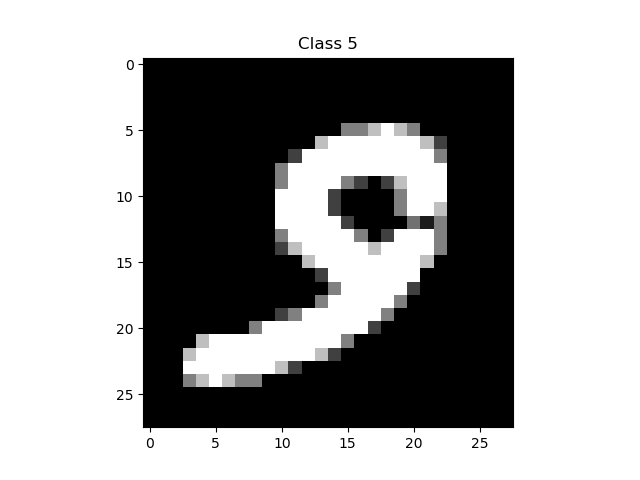}
\\
\multicolumn{1}{c}{} & \multicolumn{1}{c}{Label: Dog} & \multicolumn{1}{c}{Label: Cat} & \multicolumn{1}{c}{Label: Airplane} & \multicolumn{1}{c}{Label: 4} & \multicolumn{1}{c}{Label: 9} & \multicolumn{1}{c}{Label: 5}
\\
(d) &
\includegraphics[width=1.0in]{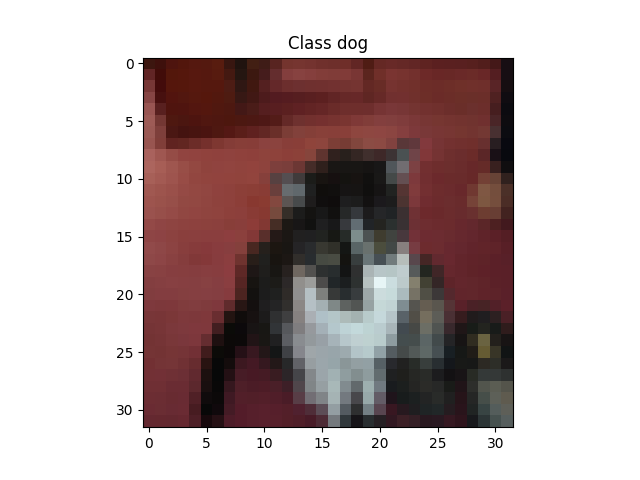} &  
\includegraphics[width=1.0in]{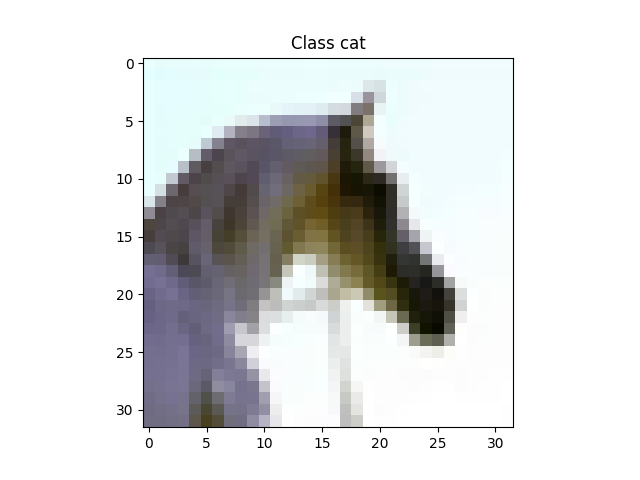} & 
\includegraphics[width=1.0in]{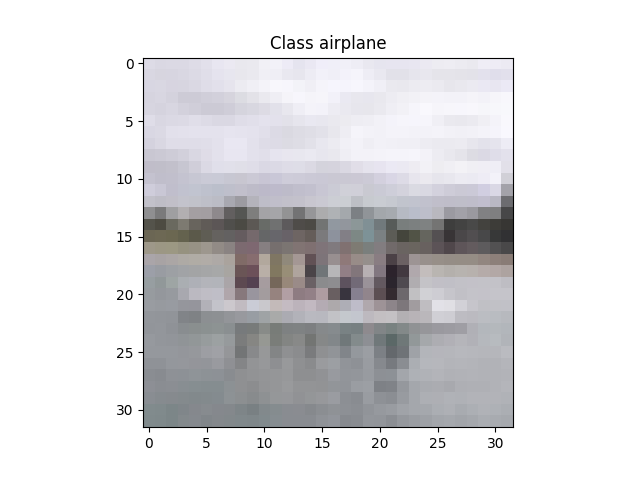} &
\includegraphics[width=1.0in]{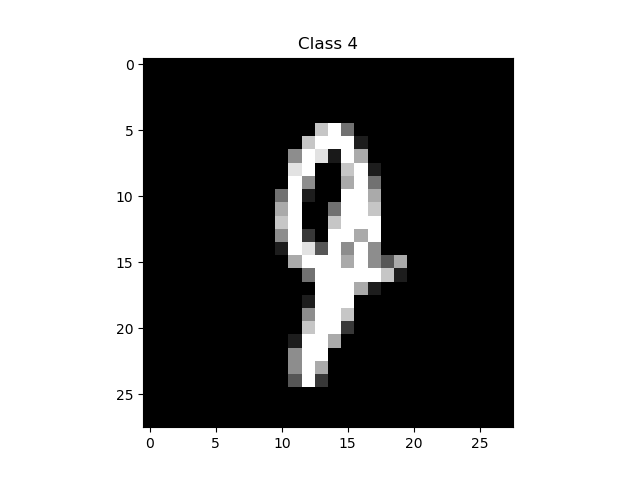} &  
\includegraphics[width=1.0in]{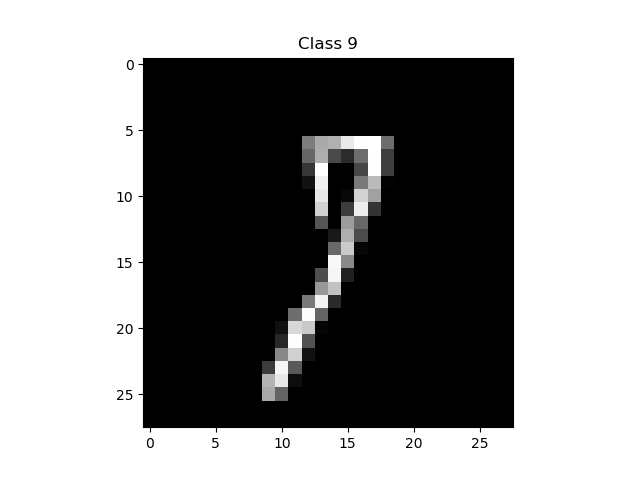} & 
\includegraphics[width=1.0in]{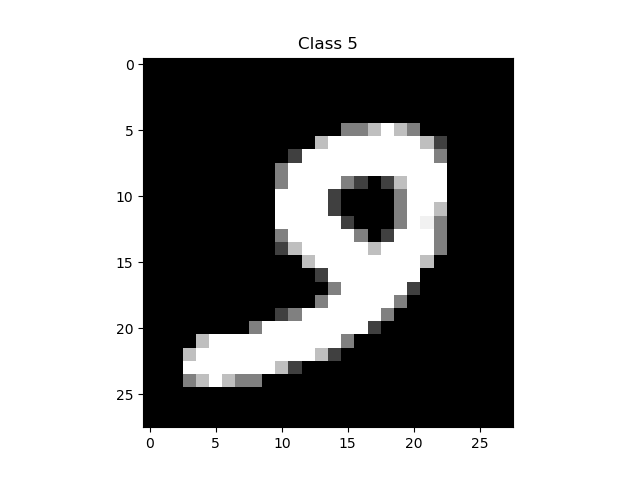}
\\
\multicolumn{1}{c}{} & \multicolumn{1}{c}{Label: Dog} & \multicolumn{1}{c}{Label: Cat} & \multicolumn{1}{c}{Label: Airplane} & \multicolumn{1}{c}{Label: 4} & \multicolumn{1}{c}{Label: 9} & \multicolumn{1}{c}{Label: 5}
\end{tabular}
\caption{Untargeted SWISS attacks on CIFAR-10 and MNIST : a) Before SWISS (b) After 1st localization (c) After complete localization (d)After final reduction}
\end{figure*}

\subsection{Hyper-parameter Analysis}\label{sec:partest}
To assess the impact of different hyper-parameter values on the two attacks, we perform empirical studies with varying swarm sizes, change rates, and $L_{\infty}$ bounds. To test the sensitivity of AdversarialPSO to the change rate hyper-parameter $m$, we fix the number of particles while modifying the change rate. We also execute the attack with fixed-sized swarms on varying change rates and $L_{\infty}$ bounds. We conduct this study by randomly selecting 100 samples from the CIFAR-10 dataset and launching the attack with the different parameter values. These tests were performed without any heuristics and without performing the reduction step, as discussed in Section III.

One advantage of using bigger swarms is the capability of synthesizing adversarial examples with smaller $L_{\infty}$ constraints. This is a natural consequence of having more particles that can evaluate an increasing number of different perturbations. This property was observed when comparing swarms of 10 and 100 particles, where 100 particle swarms were able to achieve 83\% effectiveness when setting the $L_{\infty}$ constraint to 0.075. The smaller swarm on the other hand, was not as successful, as they were only able to achieve 63\% success rates even when the $L_{\infty}$ constraint was increased to 0.15. Using bigger swarms, however, does require submitting more queries to evaluate the perturbations introduced by each particle, which would be necessary for datasets with larger dimensions.

We found the change rate to have a negative affect on the overall effectiveness of the attack. This could be attributed to the distance penalty applied to the fitness, which would hinder particles from finding better fitness positions as the $L_{2}$ distance is increasingly being penalized when more coordinates are being modified. Table \ref{AdversarialPSO parameter sensitivity test} summarizes our findings in testing the AdversarialPSO attack with different swarm sizes, change rates, and $L_{\infty}$ bounds.
\begin{table}[t]
\caption{Affects of swarm size in terms of number of particles, change rate $m$, and $L_{\infty}$ parameters on attack performance}
\label{AdversarialPSO parameter sensitivity test}
\centering
\begin{tabular}{|c|c|c|c|c|c|}
\hline
Size & $m$ & $L_{\infty}$ & Effectiveness & Avg. $L_{2}$ & Avg. Queries  
\\
\hline
10 & 0.1 & 0.15 & 63\% & 2.5697 & 215
\\
\hline
10 & 0.25 & 0.15 & 50\% & 3.7732 & 218
\\
\hline
10 & 0.5 & 0.15 & 55\% & 5.1105 & 185
\\
\hline
10 & 0.75 & 0.15 & 51\% & 5.7069 & 189
\\
\hline
10 & 1.0 & 0.15 & 43\% & 5.6551 & 166
\\
\hline
50 & 0.1 & 0.15 & 90\% & 2.4196 & 912
\\
\hline
100 & 0.1 & 0.15 & 96\% & 2.3964 & 1618
\\
\hline
100 & 0.1 & 0.1 & 90\% & 1.7179 & 2184
\\
\hline
100 & 0.1 & 0.075 & 83\% & 1.2961 & 2764
\\
\hline
\end{tabular}
\end{table}

\subsection{Observations}
Considering that we only use the velocity vector to determine the direction of the perturbations, the distance between a particle's current position and either it's best position or the swarm's best position will only have an affect if they are large enough to affect the sign of the outcome. Otherwise, the C1 and C2 weights play a more significant role if the distances towards the particle's best and the swarm best are somewhat similar in magnitude. To alleviate this imbalance, where the distance towards the swarm best will have a higher chance to be greater in earlier iterations and the distance towards the particle best will be greater in later iterations, we found that having dynamic C1 and C2 weights produce overall better results. In other words, by starting the PSO search with higher C2 weights that are reduced with every iteration and C1 weights that start low and are increased with every iteration, the swarm will have an overall better performance. This was observed on the CIFAR-10 dataset, as is has higher dimensionality than the MNIST dataset. 

For the SWISS attack, setting the change rate, step size, and $L_{\infty}$ bound to 1.0 produced the best results. The attack was more successful in generating adversarial examples with fewer queries when using fewer particles, which translates to larger chunks of the search space being assigned to each particle. This occurs, however, at the expense of larger $L_2$ distances. Essentially, each particle would expand the perturbed region to the boundaries of the search space and then check for adversarial examples. The region is then reduced in every iteration until no additional reductions are possible without losing the adversarial examples that were previously found.

Our initial reduction process would directly reduce all the coordinates that were perturbed when creating the adversarial example. However, we found that dropping the excess coordinates first before reducing the distance decreased the average number of queries by almost a third. We do that by iterating through all the coordinates that were perturbed and completely removing the part of the region due to perturbing each coordinate in turn. We then evaluate the sample by querying the model. If the label reverts back to the original label, we simply reinstate the initial change, and otherwise, we continue to the next coordinate.   
\begin{table*}[t]
\caption{Results comparison: The average queries for ZOO were calculated based on the number of iterations reported by Chen et al.~\cite{Chen:2017} times the number of gradient evaluations per iteration. For the C\&w Blackbox attack, the average time is the time required to train the surrogate and to generate the adversarial example. Furthermore, the number of queries for C\&W is the number of queries needed to train the surrogate and was calculated based on the using 150 images and five Jacobian augmentation epochs as used by Chen et al. in their experiments~\cite{Chen:2017}}
\label{AdversariaPSO and SWISS attack comparison against ZOO and C&W attacks}
\centering
\begin{tabular}{|c|c|c|c|c|c|c|c|c|}
\hline
\multicolumn{9}{|c|}{MNIST}\\
\hline
\multicolumn{1}{|c|}{} & \multicolumn{4}{c|}{Untargeted} & \multicolumn{4}{c|}{Targeted}\\
\hline
 Attack & Success Rate & Avg. $L_{2}$ & Avg. Time & Avg. Queries & Success Rate & Avg. $L_{2}$ & Avg. Time & Avg. Queries\\
 \hline
 ZOO & \textbf{100\%} & \textbf{1.49550} & 1.38 min & 384,000 & \textbf{98.9\%} & \textbf{1.987068} & 1.62 min & 384,000 \\
 \hline
 C\&W Blackbox & 33.3\% & 3.6111 & 6.92 min & 4650 & 26.74\% & 5.272 & 6.96 mins & 4650\\
 \hline
 AdversarialPSO & 96.3\% & 4.1431 & \textbf{0.068 mins} & \textbf{593} & 72.57\% & 4.778 & \textbf{0.238 mins} & \textbf{1882}\\
 \hline
 SWISS & \textbf{100\%} & 3.4298 & 0.087 mins & 3043 & 19.41\% & 3.5916 & 0.345 mins & 20026\\
 \hline
 \multicolumn{9}{|c|}{CIFAR-10}\\
\hline
\multicolumn{1}{|c|}{} & \multicolumn{4}{c|}{Untargeted} & \multicolumn{4}{c|}{Targeted}\\
\hline
 Attack & Success Rate & Avg. $L_{2}$ & Avg. Time & Avg. Queries & Success Rate & Avg. $L_{2}$ & Avg. Time & Avg. Queries\\
 \hline
 ZOO & \textbf{100\%} & \textbf{0.19973} & 3.43 mins & 128,000 & \textbf{96.8\%} & \textbf{0.39879} & 3.95 mins & 128,000 \\
 \hline
 C\&W Blackbox & 25.3\% & 2.9708 & 8.28 mins & 4650 & 5.3 \% & 5.7439 & 8.3 mins & 4650\\
 \hline
 AdversarialPSO & 99.6\% & 1.4140 & 0.139 mins & \textbf{1224} & 71.97\% & 2.9250 & \textbf{0.6816 min} & \textbf{6512}\\
 \hline
 SWISS & 99.8\% & 2.3248 & \textbf{0.1264 mins} & 2457 & 31.93\% & 2.9972 & 1.623 mins & 45308
\\
 \hline
\end{tabular}
\end{table*}

\subsection{Results}
\subsubsection{AdversarialPSO}
As can be seen in Table \ref{AdversariaPSO and SWISS attack comparison against ZOO and C&W attacks}, the AdversarialPSO attack is successful with just 593 and 1224 queries to generate untargeted adversarial examples on MNIST and CIFAR-10, respectively. In contrast, C\&W Blackbox requires over 4,000 queries, while ZOO requires over 100,000 queries---fully two orders of magnitude more queries than AdversarialPSO. In a real-world setting where the queries could be easily monitored by MLaaS providers and other remote hosts, submitting such a large number of queries would be impractical. AdversarialPSO also runs much faster than either attack in just seconds instead of minutes.

AdversarialPSO also offers very high success rates of 96.3\% and 99.6\% on MNIST and CIFAR-10, respectively. This is almost as good as the ZOO attack and much better than the 33.3\% and 25\% respectively achieved by the C\&W Blackbox attack. The average $L_{2}$ distance between the inputs and the adversarial examples is reasonable at 4.1 and 1.4, respectively. These distances are higher than those generated by ZOO. Examples of untargeted adversarial examples generated using AdversarialPSO on CIFAR-10 and MNIST can be seen in Fig. 1, and we observe minimal visually recognizable differences between the inputs and corresponding adversarial examples. Notably, the visible changes do not seem suspicious and do not show signs to suggest that the images have been manipulated when seen on their own instead of compared to the originals. Given this, we believe that the AdversarialPSO outputs are sufficiently convincing at these $L_2$ distances.

For targeted attacks, AdversarialPSO also requires much fewer queries than ZOO at 1882 and 6512 for MNIST and CIFAR-10, respectively, again compared to over 100,000 for ZOO. AdversarialPSO is, however, less successful than ZOO at finding these examples, with 72.57\% success rate for MNIST and 71.97\% for CIFAR-10. 
The same parameters used to launch the untargeted attacks, were used for targeted attacks. However, we removed the early termination condition and allowed the search process to run for a maximum of 300 iterations.

\subsubsection{SWISS}
The SWISS attack was also able to generate adversarial examples. By using 3 particles to perform untargeted attacks, adversarial examples were generated by perturbing large portions of the input to the boundaries of the data manifold and then reducing. This, however, generally required more queries than AdversarialPSO and would create adversarial examples with higher $L_{2}$ distances. MNIST was the only exception with regards to $L_{2}$ as SWISS, on average, generated adversarial examples closer to their sources than AdverarialPSO. The average $L_{2}$ increased with higher dimension datasets, However, this could be balanced by increasing the number of particles, which would raise the precision of the attack and create adversarial examples with lower $L_{2}'s$, but with more queries. Examples of images generated using the SWISS attack can be seen in Fig. 2 and the progression of the SWISS localization process from inputs to adversarial examples can be seen in Fig. 3. 

\begin{figure*}[t]
\begin{tabular}{m{1in}m{1in}m{1in}m{1in}m{1in}m{1in}}
\includegraphics[width=1.2in]{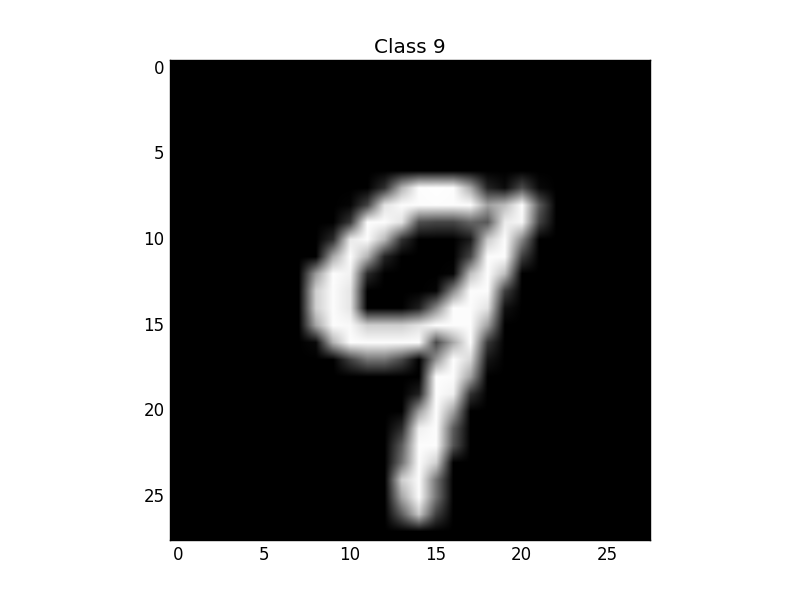} &  
\includegraphics[width=1.2in]{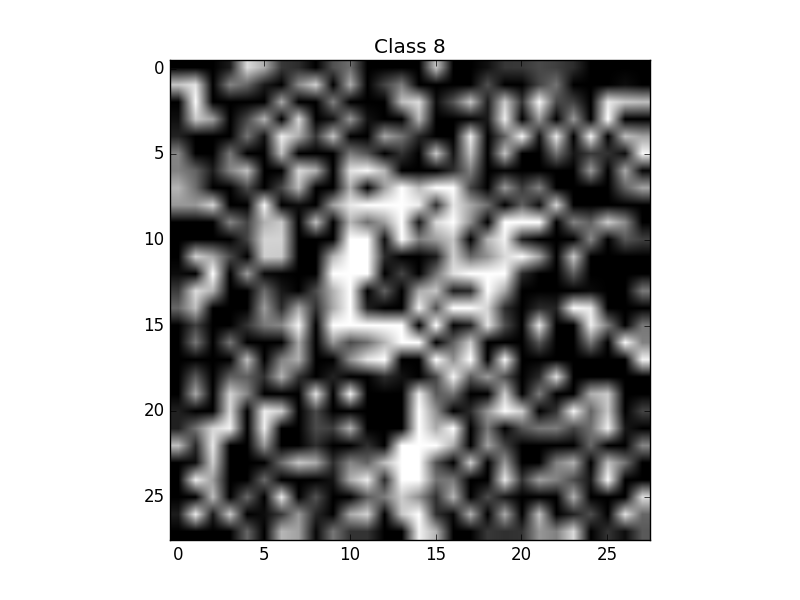} & 
\includegraphics[width=1.2in]{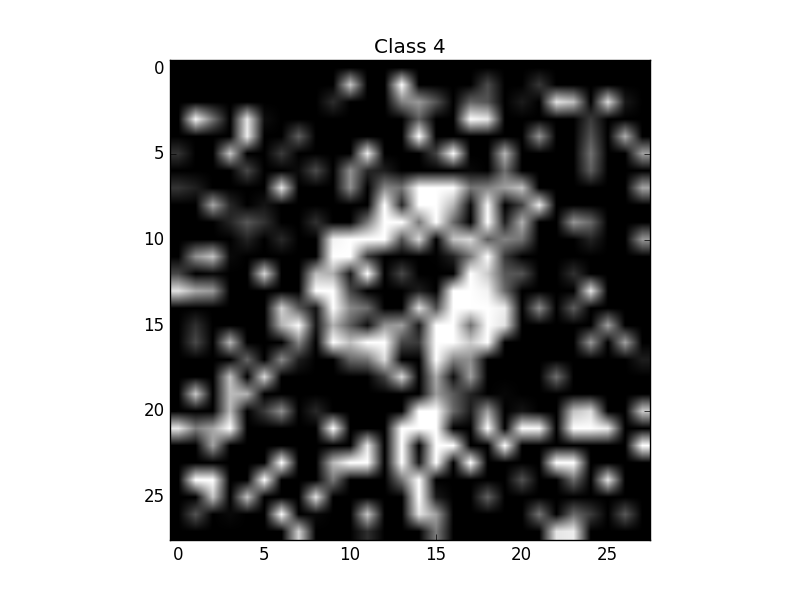} &
\includegraphics[width=1.2in]{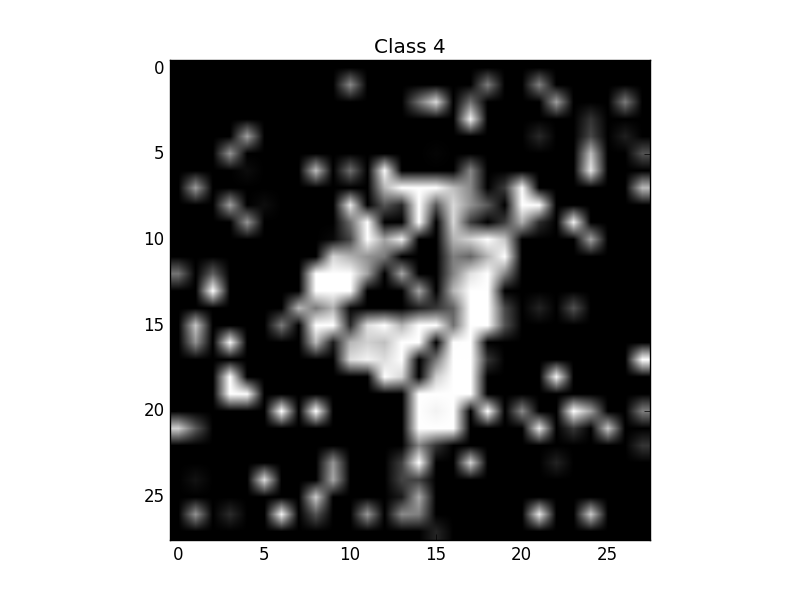} &  
\includegraphics[width=1.2in]{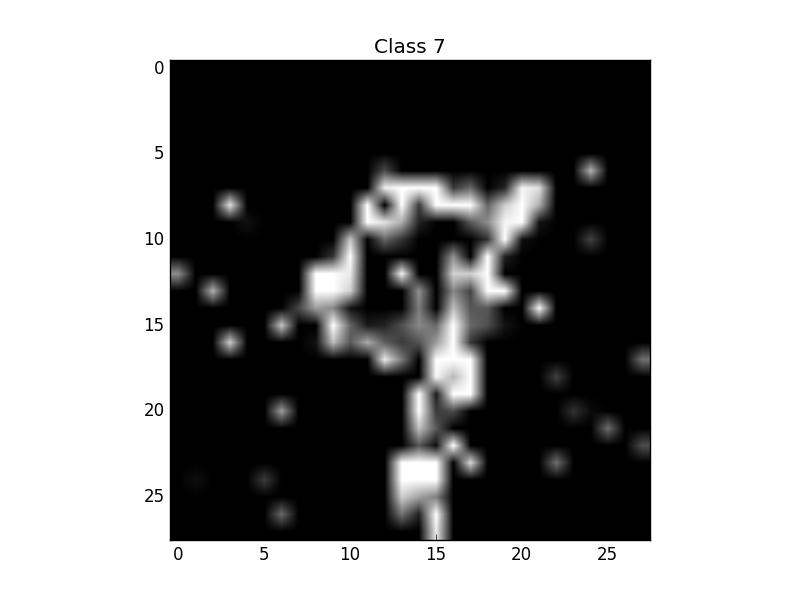} & 
\includegraphics[width=1.2in]{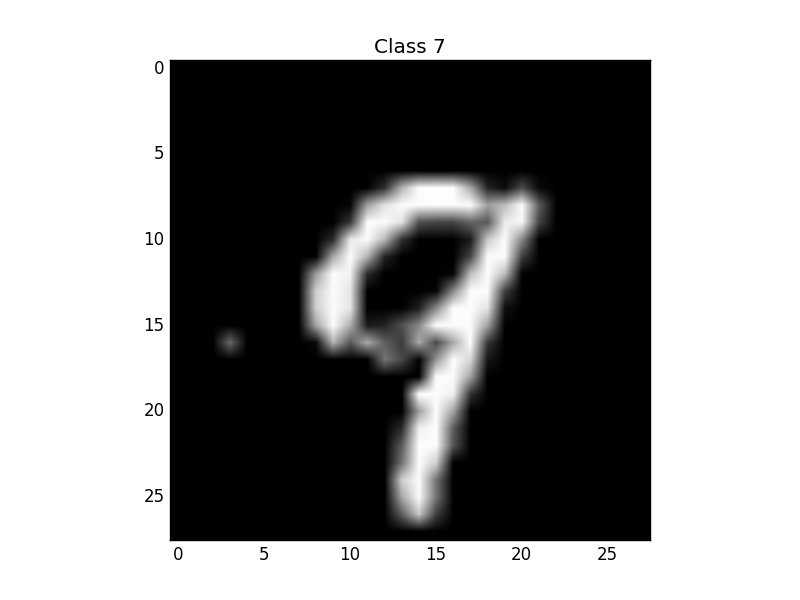} 
\\
\multicolumn{1}{c}{(a) Label: 9} & \multicolumn{1}{c}{(b) Label: 8} & \multicolumn{1}{c}{(c) Label: 4} & \multicolumn{1}{c}{(d) Label: 4} & \multicolumn{1}{c}{(e) Label: 7} & \multicolumn{1}{c}{(f) Label: 7}
\\
\includegraphics[width=1.2in]{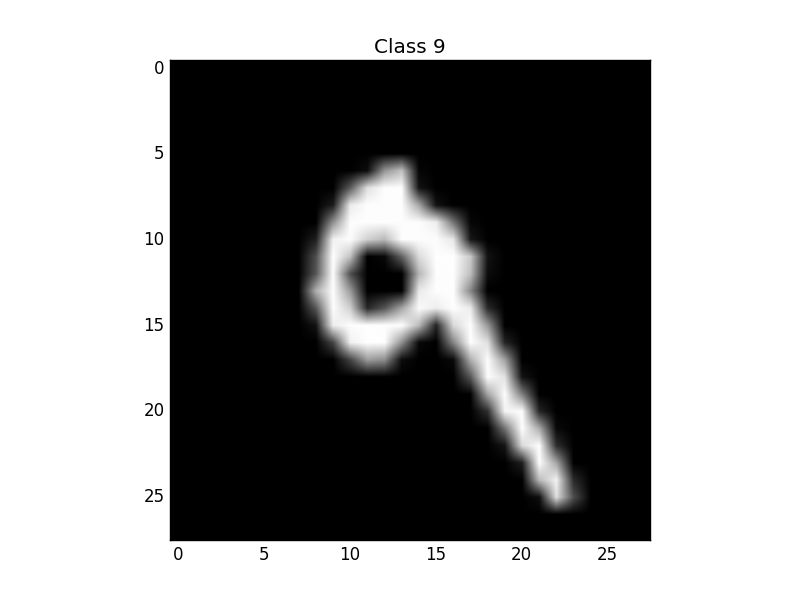} &  
\includegraphics[width=1.2in]{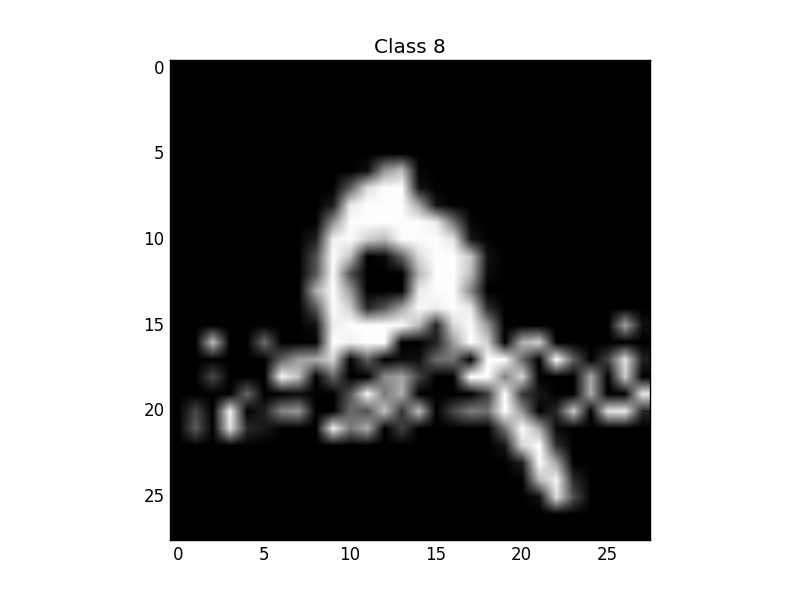} & 
\includegraphics[width=1.2in]{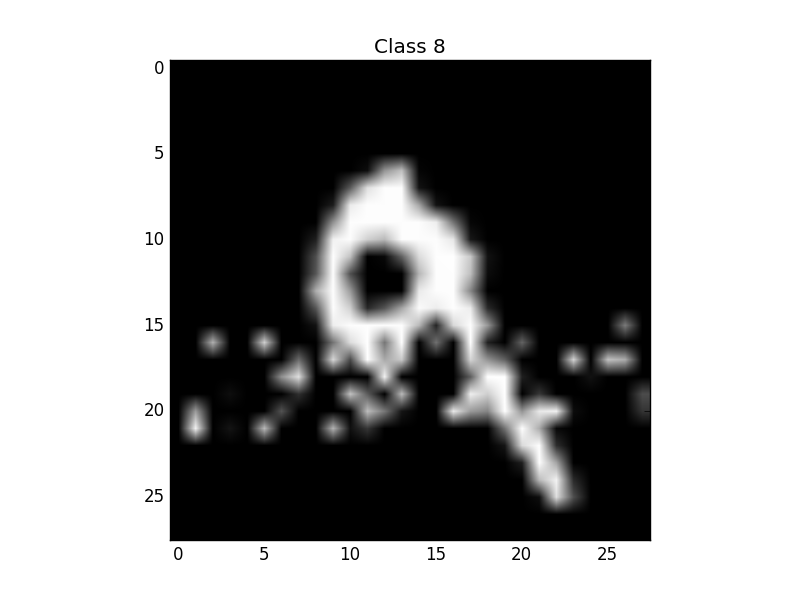} &
\includegraphics[width=1.2in]{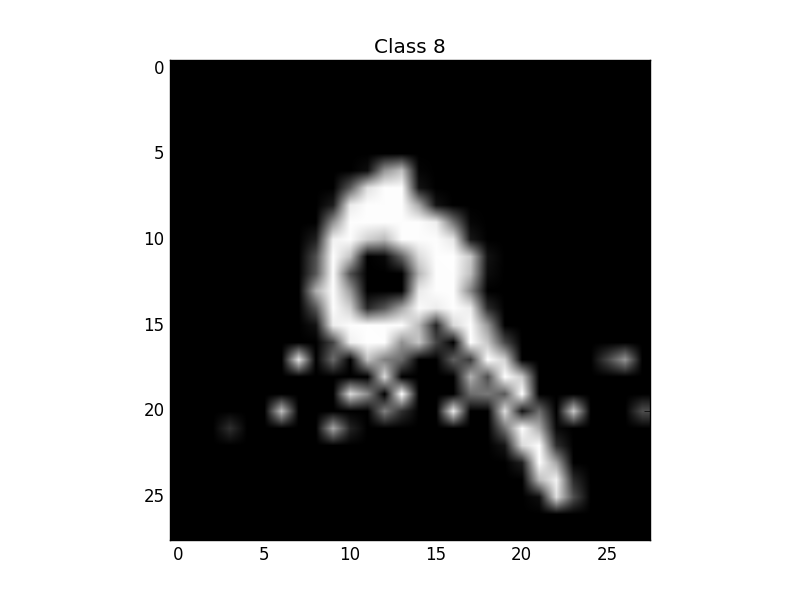} &  
\includegraphics[width=1.2in]{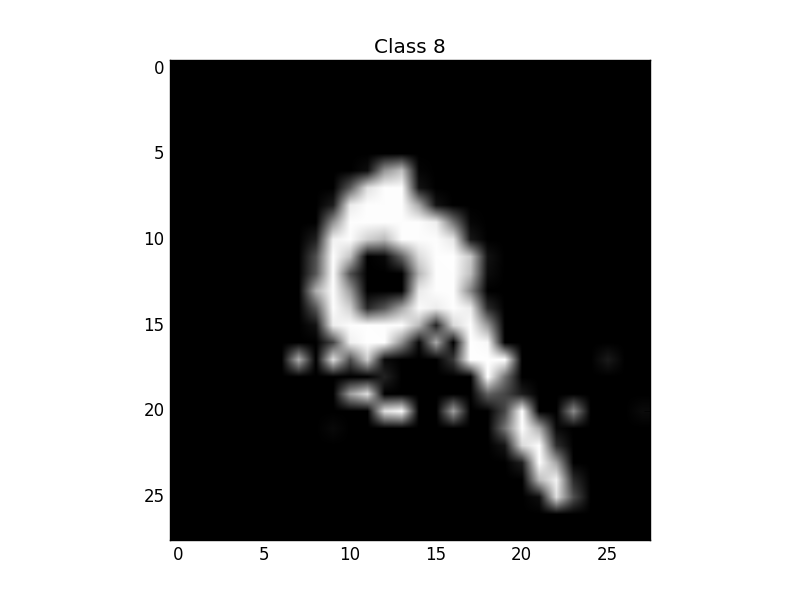} & 
\includegraphics[width=1.2in]{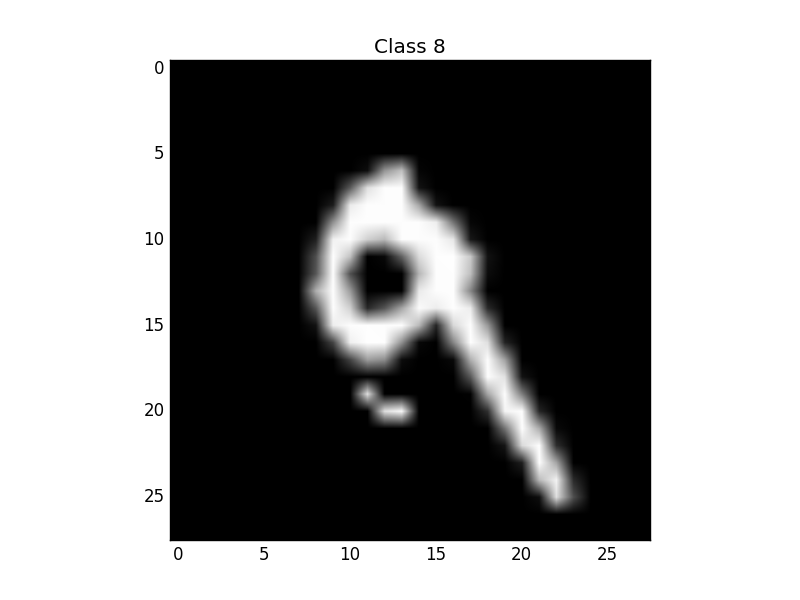} 
\\
\multicolumn{1}{c}{(a) Label: 9} & \multicolumn{1}{c}{(b) Label: 8} & \multicolumn{1}{c}{(c) Label: 8} & \multicolumn{1}{c}{(d) Label: 8} & \multicolumn{1}{c}{(e) Label: 8} & \multicolumn{1}{c}{(f) Label: 8}
\end{tabular}
\caption{Progression of SWISS attack localization from start to adversarial example: (a) Before SWISS attack (b) After 1st localization (c) After 2nd localization (d) After 3rd localization (e) After 4th localization (f) After final reduction}
\end{figure*}
\subsection{High-Dimensional Evaluation on Imagenet}
To evaluate the performance of AdversarialPSO on high-dimensional datasets, we launched the attack on InceptionV3 using the first 150 correctly classified test samples of the Imagenet dataset. By only utilizing a single CPU core to calculate particle movements, the attack did require more time to find adversarial examples, where the average run-time per sample was 5.1 minutes using a swarm of only 20 particles. As shown in Table \ref{AdversariaPSO on Imagenet}, although the average $L_{2}$ distance between Imagenet samples and their adversarial counterparts is 3 times larger than ZOO, the number of queries used to generate those adversarial examples is approximately 450 times less. Furthermore, the AdversarialPSO attack successfully generated adversarial examples for 82\% of the images. Examples of the generated adversarial examples on the Imagenet dataset could be seen in Fig. 4.
\begin{table}[t]
\caption{AdversarialPSO performance on Imagenet}
\label{AdversariaPSO on Imagenet}
\centering
\begin{tabular}{|c|c|c|c|}
\hline
Attack & Success Rate & Avg. $L_{2}$ & Avg. Queries
\\
\hline
ZOO & \textbf{88.9\%} & \textbf{1.19916} & 1,280,000
\\
\hline
AdversarialPSO & 82.00\% & 3.8304 & \textbf{2833}
\\
\hline
SWISS & 70.66\% & 4.9331 & 8429
\\
\hline
\end{tabular}
\end{table}

In comparison to the GenAttack, which is summarized in Table~\ref{AdversarialPSO and GenAttack comparison}, AdversarialPSO performed better in either the number of queries, the quality of the adversarial images, or in both. Results are obtained by running the authors code with the recommended parameters.\footnote{\url{https://github.com/nesl/adversarial_genattack}} The only parameter we modified in the maximum steps for Imagenet, where the authors recommend 100,000 steps, we set it to 5,000. With a population size of six, this configuration allows a maximum of 30,000 queries. We did so to evaluate the attack with a limited number of queries to create a more realistic setting. Without this limit, the authors report 100\% effectiveness with an average of 11,081 queries and 7.8540 $L_{2}$ distance. As the attack is allowed to run for more steps, many of the adversarial examples were created after submitting a high number of queries and adding large perturbations to the input, which explains the higher query count and larger $L_{2}$ average.

On MNIST and Imagenet, AdversarialPSO generated better quality adversarial examples using a third of the queries used by GenAttack. On CIFAR-10, although the average $L_{2}$ produced by AdversarialPSO was 3\% more than that produced by GenAttack, the average number of queries used by AdversarialPSO is 10\% less. 

\begin{table}[t]
\caption{AdversarialPSO and GenAttack comparison}
\label{AdversarialPSO and GenAttack comparison}
\centering
\begin{tabular}{|c|c|c|c|}
\hline
\multicolumn{4}{|c|}{AdversarialPSO}\\
\hline
Dataset & Success Rate & Avg. $L_{2}$ & Avg. Queries
\\
\hline
MNIST & \textbf{96.3}\% & \textbf{4.1431} & \textbf{593}
\\
\hline
CIFAR-10 & \textbf{99.6\%} & 1.414 & \textbf{1224}
\\
\hline
Imagenet & \textbf{82.00\%} & \textbf{3.8304} & \textbf{2833}
\\
\hline
\multicolumn{4}{|c|}{GenAttack}\\
\hline
Dataset & Success Rate & Avg. $L_{2}$ & Avg. Queries
\\
\hline
MNIST & 94.45\% & 5.1911 & 1801
\\
\hline
CIFAR-10 & 98.09\% & \textbf{1.3651} & 1360
\\
\hline
Imagenet & 74.66\% & 4.5563 & 8590
\\
\hline
\end{tabular}
\end{table}

\subsection{Wide Applicability of AdversarialPSO and SWISS attacks}
To test the applicability of the two attacks, we evaluate their effectiveness on two additional models for each dataset. For CIFAR-10, we evaluate the attacks on the CNN-Capsule model trained with data augmentation as described by Sabour et al.~\cite{Sabour:2017} and the ResNet model proposed by He et al.~\cite{He:2015}. These target models achieve a test set accuracy of 82.43\% and 85.71\%, respectively. For MNIST, we train a Hierarchical Recurrent Neural Network~\cite{Li:2015} with an accuracy of 98.46\% and a simple Multi-Layer Perceptron with 98.40\% accuracy. Table~\ref{AdversariaPSO and SWISS attacks on additional models} summarizes the results of this test. As shown in the table, both attacks successfully generated adversarial examples on all target models, including the CNN-Capsule model, which uses augmented samples in the training process. This shows the wide applicability of the attacks on different models. 

\begin{table*}[t]
\caption{Wide-applicability results of AdversarialPSO and SWISS}
\label{AdversariaPSO and SWISS attacks on additional models}
\centering
\begin{tabular}{|c|c|c|c|c|c|c|}
\hline
\multicolumn{7}{|c|}{MNIST}\\
\hline
\multicolumn{1}{|c|}{} & \multicolumn{3}{c|}{HRNN} & \multicolumn{3}{c|}{MLP}\\
\hline
 Attack & Success Rate & Avg. $L_{2}$ & Avg. Queries & Success Rate & Avg. $L_{2}$  & Avg. Queries\\
 \hline
 AdversarialPSO & 100.0\% & 2.4497 & 552 & 94.7\% & 3.1831 & 548 \\
 \hline
 SWISS & 100.0\% & 2.5136 & 3214 & 100.0\% & 3.2096 & 1984\\
 \hline
 \multicolumn{7}{|c|}{CIFAR-10}\\
\hline
\multicolumn{1}{|c|}{} & \multicolumn{3}{c|}{CNN\_Capsule} & \multicolumn{3}{c|}{ResNet}\\
\hline
 Attack & Success Rate & Avg. $L_{2}$ & Avg. Queries & Success Rate & Avg. $L_{2}$ & Avg. Queries\\
 \hline
 AdversarialPSO & 97.8\% & 1.3343 & 2052 & 100.0\% & 0.5774 & 1723 \\
 \hline
 SWISS & 98.9\% & 2.0051 & 3725 & 100.0\% & 1.8391 & 1792\\
 \hline
\end{tabular}
\end{table*}
 
\section{Discussion}
\subsection{Potential Improvements}
Our current implementation of the AdversarialPSO and SWISS attacks execute sequentially, where particles are looped through and moved one by one. This design choice was made to evaluate the speed of the attacks with the least amount of resources. A natural next step is to parallelize the process by dividing the particles among multiple CPU or GPU cores, allowing for faster execution of bigger swarms. It would not, however, affect either the $L_{2}$ difference between the adversarial examples and their inputs nor the number of queries submitted to the target if the swarm size remains unchanged.

An extension to PSO is Multi-Swarm Optimization (MSO), which uses multiple sub-swarms instead of a single swarm. Using MSO with different starting points could help cover more areas of the search space and eventually find better adversarial examples. This, however, would add to the total number of queries if each MSO swarm is the same size as the PSO swarm. To avoid this, smaller sub-swarms can be used in MSO so that the total number of particles remains unchanged.

Having multiple sub-swarms also allows the use of different fitness functions or different swarm configurations. The sub-swarms could apply different distance penalties, mutate particles more or mutate them less, have different step sizes, or have different exploitation/exploration/inertia weights. Sharing the same starting point but with different configurations could allow better exploration of different regions of the search space.

\subsection{Limitations of Gradient-Based Attacks}
Launching gradient-based black-box attacks could be made with or without using a local surrogate; both approaches have their limitations. First and foremost, when using a local surrogate that approximates the target, the success of the attack depends on the precision and efficacy of several intermediary steps that could affect the overall attack. The attacker must have a tight approximation of the target trained locally, and the adversarial examples must successfully transfer from the surrogate to the remote target. As we have seen in Section IV, the C\&W attack, which is capable of generating high-quality adversarial examples in a white-box setting, achieved low success rates when training on a surrogate in a black-box setting. This could be attributed to a poorly trained local surrogate that produces adversarial examples incapable of transferring to the remote target. 

Alternatively, a gradient-based black-box attack could be launched directly on the target by estimating the gradients as was done by the ZOO attack. Although the attack does generate high-quality adversarial examples, it requires a large number of queries to do so. Submitting multitudes of queries to the target is impractical in a real-world attack, as that would easily be noticed by an operator that is monitoring the volume of incoming queries.

Although gradient-based attacks have been proven to generate high-quality adversarial examples, we argue that they are impractical in their current state when launching black-box attacks. Gradient-based techniques are finely tuned to eventually finding global optima, which simply requires more queries to be successful. The PSO approach thus appears to offer a better trade-off in this setting, where it more quickly converges to high-quality but non-optimal examples. Considering that a black-box scenario is a more likely setting, particularly in security-sensitive settings, attacks must be able to generate adversarial examples with a realistic number of queries. 

\subsection{Adversarial Examples in the Malware Domain}
While images are an interesting test case, evading detection algorithms is a more clearly security-specific setting where adversarial examples are important to understand. AdversarialPSO could be adapted to generate adversarial examples in the malware domain, for example. By considering malware as an $n$-dimensional discrete vector, where each element is an API call used by the malware, this problem could be treated as a combinatorial optimization problem, an area where PSO has previously had success~\cite{Chen:2010}. As many API calls have alternatives that perform the same functionality, those calls could be swapped for their alternatives without changing the malware's behavior. The challenge would be to find those alternatives and adjust the necessary parameters. Furthermore, the changes must be reflected in malware code without breaking the malware functionality by introducing syntax or run-time errors. This presents an interesting potential application of AdversarialPSO or SWISS, as the adversary may well be attempting to evade a black-box classifier that carefully limits the number of querying attempts for security reasons.

\subsection{Large-Scale Adversarial Attacks}
The AdversarialPSO attack we present in this paper provides the opportunity for attackers to launch large-scale attacks on remote targets that host models for high-dimensional datasets. Due to the nature of the attack, it can be easily scaled not only on a single machine by increasing particles, but across multiple machines that share a common target. We anticipate future attacks on machine learning models not to be launched from a single source, but from multiple sources working in conjunction. In a sense, this is similar to botnets, where a large group of zombie bots are utilized for a single purpose. The AdversarialPSO attack provides the foundation for such attacks, where tens of thousands of particles can be launched by multiple attackers. Note that PSO does not require a GPU for efficient computation, making it more suitable for this setting than gradient-based techniques, since the capabilities of the bots are more likely to be limited.

As we have seen in our experiments, using bigger swarms provides the capacity to generate adversarial examples with shorter $L_{2}$ distances. Therefore, by pooling resources together, a group of attackers can launch more powerful attacks that can generate adversarial examples that are increasingly harder to distinguish from their respective inputs. Such an attack could be scaled to accommodate inputs with higher dimensions and produce adversarial examples on models that are currently difficult to attack, such as those that are trained on 8K resolution images.  

Additionally, as the bot nodes would likely be located in different geographical locations, monitoring incoming queries would be more difficult than if the attack is originating from a single host. To complicate the attacks even further, decoy queries could be launched to confuse the target as to which entities are part of the attack. This could also be achieved by controlling the rate of queries being submitted by any single source, where queries are to be submitted in a certain order, at different times, or in random bursts.

The SWISS attack also has the potential for a large-scale distributed attack. This would be especially true for high-dimensional datasets, where each entity participating in the attack could be assigned a specific region of the input. On 8K images, which contain up to 100 million coordinates per image, attacks would require a massive amount of resources if performed by a single entity. However, by dividing the image among multiple parties, where each party is responsible for a certain region of the image, generating adversarial examples would be feasible. 
\begin{figure}[!t]
\begin{tabular}{lm{1in}m{1in}}
(a)  &
\includegraphics[width=1.0in]{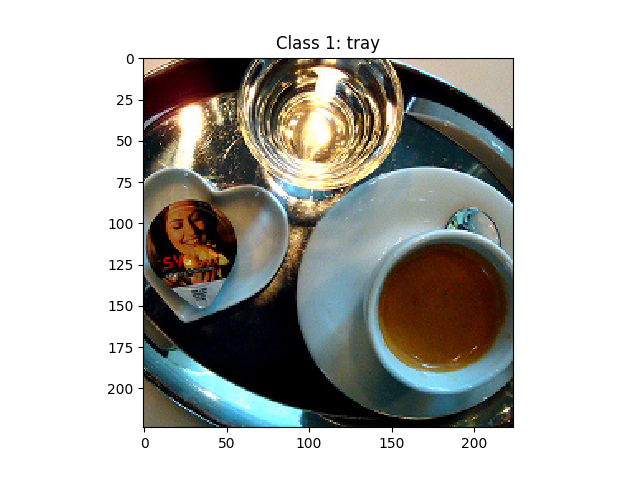} &  
\includegraphics[width=1.0in]{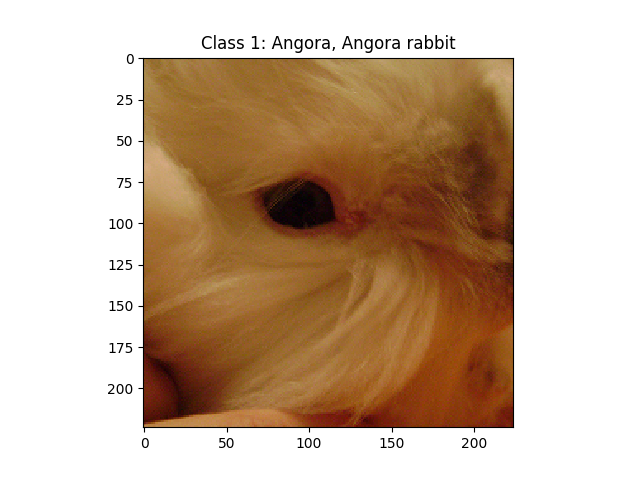} 
\\
\multicolumn{1}{c}{} & \multicolumn{1}{c}{Label: Tray} & \multicolumn{1}{c}{Label: Rabbit}
 \\
(b) &
\includegraphics[width=1.0in]{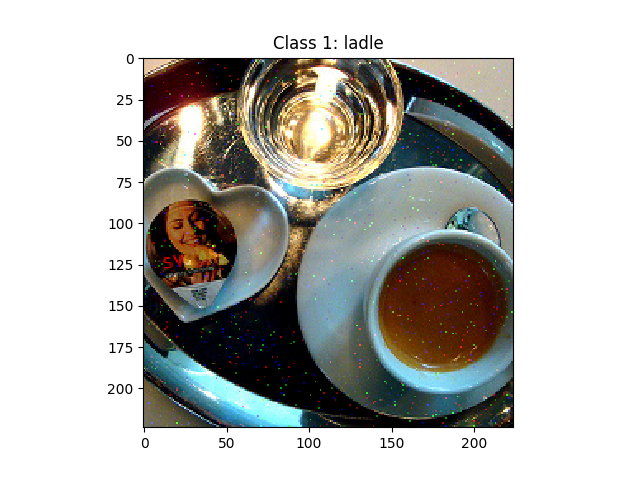} & \includegraphics[width=1.0in]{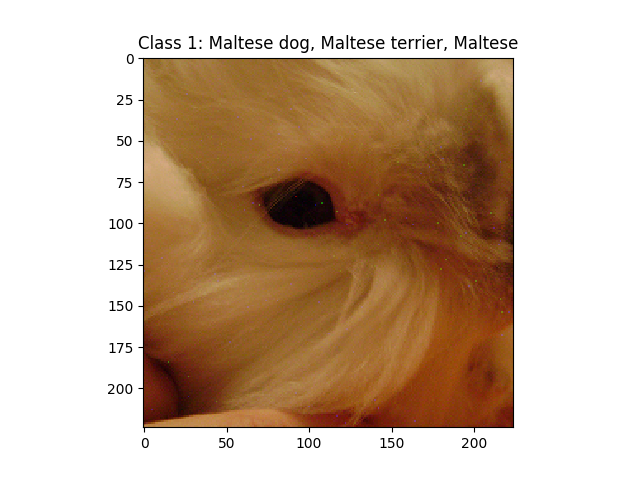}
\\
\multicolumn{1}{c}{} & \multicolumn{1}{c}{Label: Ladle} & \multicolumn{1}{c}{Label: Dog}
 \\
(c) &
\includegraphics[width=1.0in]{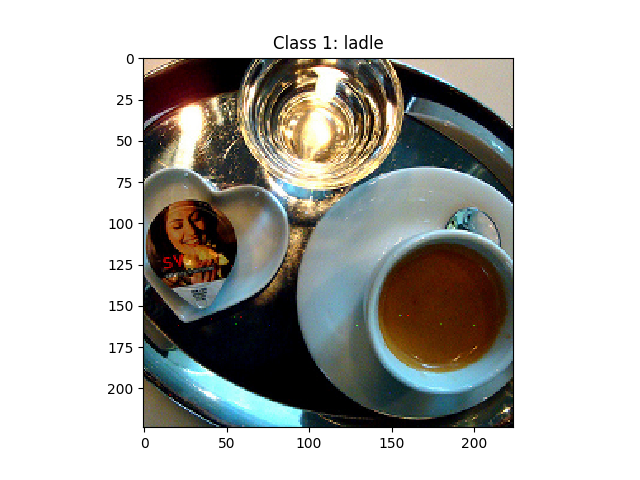} &
\includegraphics[width=1.0in]{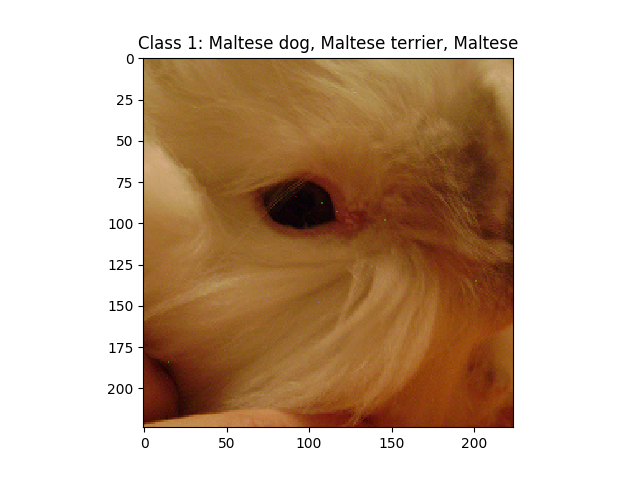}
\\
\multicolumn{1}{c}{} & \multicolumn{1}{c}{Label: Ladle} & \multicolumn{1}{c}{Label:  Dog} 
\end{tabular}
\caption{Untargeted AdversarialPSO attacks on Imagenet : a) Before AdversarialPSO (b) After AdversarialPSO but before reduction (c) After reduction}
\end{figure}
\section{Conclusions}
This paper presented two black-box attacks based on the evolutionary search algorithm Particle Swarm Optimization. The first attack adapts the traditional PSO algorithm to produce adversarial examples from images. The second utilizes the PSO infrastructure to launch a localization-based attack that finds salient features used by the target model for classification. Our experimental evaluations on the MNIST, CIFAR10, and Imagenet datasets suggest that both attacks can effectively generate adversarial examples in practical black-box settings with a limited number of queries to the target model. The purpose of these two attacks is to help evaluate security-critical models against black-box attacks and to promote the search for robust defenses.

\section*{Acknowledgment}
This material is based upon work supported by the National Science Foundation under Awards No. 1816851 and 1433736.
\bibliographystyle{IEEEtran}
\bibliography{references}

\end{document}